\documentclass[12pt]{article}
\pdfoutput=1
\usepackage{eurosym}
\usepackage[totalwidth=460truept,totalheight=620truept,heightrounded]{geometry}
\usepackage{latexsym,graphicx,amsfonts,amssymb,amsmath,xcolor}
\usepackage[hypertexnames=false,hidelinks]{hyperref}
\usepackage{fancyhdr}
\usepackage{booktabs}

\def\theequation{\arabic{section}.\arabic{equation}}

\renewcommand{\theequation}{\thesection.\arabic{equation}}
\global\arraycolsep=1truept
\renewcommand{\theequation}{\arabic{section}.\arabic{equation}}

\begin{document}

\null

\begin{center}
{\huge \textbf{Exact Solutions and}}

\vskip.55truecm

{\huge \textbf{Topological Phase Transitions}}

\vskip.5truecm

{\huge \textbf{to Boolean Logic in Minimal RNNs}}

\vskip1truecm

\textsl{Damiano Anselmi}

\url{https://renormalization.com}

\href{mailto:damiano.anselmi@renormalization.com}{damiano.anselmi@renormalization.com}

\vskip1truecm

\textbf{Abstract}
\end{center}

\vskip-.1truecm 

The mechanistic interpretability of Recurrent Neural
Networks (RNNs) executing exact algorithmic tasks remains challenging, as
continuous latent spaces typically entangle logic within approximate,
non-convex manifolds. We study continuous binary addition with carry using a
minimal RNN with hyperbolic tangent activation. Introducing a saturated
penalty---the repulsive dead-zone loss---on datasets of 256, 64, and even 16
additions, we show that the optimization landscape abruptly undergoes a
topological phase transition to Boolean logic. Rather than stagnating in
local minima, this transition quantizes the dynamics into compact,
infinitely stable logical regions separated by a finite gap and mapped into
their interiors, guaranteeing zero error over arbitrary sequence lengths. We
prove exact invariance and zero-error execution in latent dimensions $d=2,3,4$.
Crucially, the domain of structurally stable exact solutions includes
nonempty open subsets in parameter space encompassing a dense set of dyadic
rationals, ensuring that an appropriately optimized RNN inevitably converges
into error-absorbing logical regions with infinite zero-error stability
without requiring infinite precision. We further demonstrate hierarchical
composability by using the exact neural adder as an immutable Arithmetic
Logic Unit directed by a lightweight, memoryless neural controller to
orchestrate sequential shift-and-add multiplication with absolute zero error
over arbitrary lengths. Decomposing complex tasks into a hierarchy of
topologically protected, statistically certain micro-tasks thus enables
continuous dynamical systems to construct exact logical scaffolding. Serving
as a proof of concept for the verticalization of AI via structurally stable
neural modules, this result provides a robust path toward a
``Super AI'' capable of infinite-horizon
reasoning.

\vfill\eject

\section{Introduction}

\label{sec:introduction}

The emergence of deep neural network architectures has revolutionized
machine learning, demonstrating extraordinary capabilities across a vast
array of complex tasks. However, the inherently opaque nature of these
models---often treated as statistical black boxes---represents a primary
obstacle to fundamental understanding and deployment in contexts requiring
formal rigor and verification. Within this landscape, Recurrent Neural
Networks (RNNs) provide an ideal proving ground for mechanistic
interpretability, as their architecture forces the learning of iterative
algorithms through the evolution of a dynamical system within a continuous
phase space, known as the latent space.

Understanding how a neural network encodes exact logical operations within a
continuous and differentiable dynamic is a central open challenge. While
recent advances in mechanistic interpretability have made significant
strides in post-hoc circuit extraction and discovering internal
representations in trained models \cite{nanda,merrill}, bridging the gap
between continuous optimization and exact, provable symbolic execution
remains difficult.

In this work, we address this issue by studying one of the most fundamental
sequential algorithmic tasks: continuous binary addition with carry. This
problem requires the network not only to perform non-linear logical
operations on input bits but also to maintain persistent, long-term memory
(the logical carry) utilizing solely the geometry of its latent space and a
non-linear activation function, such as the hyperbolic tangent.

Typically, numerical training on such rigid algorithmic tasks via gradient
descent yields approximate and unstructured solutions, trapped in local
minima of a non-convex manifold. We demonstrate that by altering the
optimization landscape through a custom cost function with a saturated
penalty---which we define as the repulsive dead-zone loss---the system
escapes the stagnation characteristic of continuous regimes. Instead,
training exhibits an abrupt topological phase transition to Boolean logic:
the loss suddenly collapses to absolute zero, indicating that the network's
dynamics have quantized, locking into a globally stable geometric attractor.
This transition does not require a massive dataset to occur during training:
we observed it even with datasets containing as few as 16 additions (the
sums of the binary numbers 0, 1, 2, and 3).

While theoretical works previously established that recurrent neural
networks can formally emulate finite-state automata and Turing machines,
these classical proofs largely relied on idealizing assumptions---such as
discontinuous step activations, manually hard-coded high-magnitude weights
in second-order networks, asymptotic weight limits, or unphysical
infinite-precision irrational weights \cite%
{Giles1992,Giles1996,Siegelmann1995,Casey1996}. The framework presented here
bridges the historical divide between theoretical expressivity and practical
optimization. Rather than requiring infinite precision or manual heuristic
design, we demonstrate that training under a repulsive dead-zone loss
function triggers a spontaneous global bifurcation toward exact symbolic
computation. This mechanism fractures the continuous phase space into
robust, invariant basins of attraction that embed compact logical zones
separated by a finite gap and mapped into their interiors with a positive
safety margin, shielding internal states against numerical noise and locking
small, real-valued weight matrices into an exact, zero-error finite-state
machine under standard float32 arithmetic.

We formalize these results through the following theoretical advancements:

\begin{enumerate}
\item \textbf{Topological Phase Transitions and Exact Solutions in }$d=2$, $%
3 $\textbf{, and }$4$\textbf{:} Moving beyond empirical observation, we
provide rigorous analytical proofs of topological invariance across all
considered latent space dimensions, demonstrating that the network perfectly
embeds zero-error discrete logic. In each case, we identify invariant basins
of attraction that permanently trap the dynamic states within compact
logical zones separated by a finite gap and mapped into their interiors,
proving that the continuous dynamics collapse onto structurally stable
finite-state machines. While the weight matrices for the $\allowbreak d=2$
minimal adder were empirically discovered by the neural network itself,
guided by the repulsive dead-zone loss function---and would be highly
non-trivial to infer a priori---arguments based on symmetries enable the
direct construction of exact, closed-form analytical solutions in higher
dimensions ($d=3$ and $d=4$). Typically, however, these mathematically
elegant configurations either rely on irrational weights or exhibit
invariant basins that lack a finite topological gap. By algebraically
twisting specific weight parameters to break the symmetries, we successfully
establish a finite separation between the attractors, effortlessly yielding
structurally stable configurations with dyadic rational weights
(finite-precision machine numbers).

\item \textbf{Open Sets of Analytical Solutions:} A\ crucial result is that
the exact solutions executing Boolean logic outlined in this paper are
characterized by rational weights and invariant basins that strictly confine
the states within compact logical zones separated by a finite gap, which are
mapped by the dynamics into their interiors. This demonstrates that the
structurally stable exact configurations are not isolated points, but
include nonempty open subsets within the parameter space, providing the
fundamental reason why an appropriately optimized RNN inevitably converges
into these error-absorbing solutions without requiring infinite precision.

\item \textbf{Modular Neuro-Symbolic Composability:} We illustrate the
hierarchical composability of our approach by freezing the exact neural
adder to serve as a fixed Arithmetic Logic Unit (ALU) within a multi-bit
binary multiplier. Guided by a lightweight neural controller that
spontaneously collapses into a memoryless state during optimization, the
hybrid system dynamically orchestrates sequential shift-and-add logic over
arbitrary sequences with absolute zero error.
\end{enumerate}

Beyond solving specific arithmetic tasks, this modular composability
provides a foundational blueprint for what we term the \textit{%
verticalization of AI}. The current trajectory toward Artificial General
Intelligence (AGI) relies heavily on horizontal scaling, forcing monolithic
black boxes to approximate entire algorithmic mappings end-to-end. This
approach is statistically inefficient, mathematically opaque, and vulnerable
to catastrophic forgetting. Drawing inspiration from human cognitive
development---where complex operations such as multiplication are strictly
built upon the previously mastered prerequisite of addition---we argue that
building a theoretically unbounded \textquotedblleft Super
AI\textquotedblright\ requires fracturing statistically unattainable
computational goals into a vertical tree hierarchy of nested micro-tasks
that gradient descent can solve with near certainty when the optimization
landscape is sculpted by an appropriate penalty. By demonstrating that
continuous dynamical systems can autonomously prune unnecessary degrees of
freedom and synthesize topologically protected, composable modules, this
work serves as a mathematical proof of concept for the physics of
amplification towards Super AI.

The manuscript is organized as follows. Section \ref{sec:model_architecture}
defines the general architecture and latent dynamics of the minimal RNN
adder. Section \ref{sec:dead_zone_loss} formalizes the training phase by
means of repulsive dead-zone loss, and analyzes the thermodynamics of the
topological phase transition. In addition, it contextualizes our framework
within contemporary literature on algorithmic generalization and
regularization, comparing our findings with standard grokking phenomena and
conventional training pathologies, such as gradient explosion and latent
space destabilization. Section \ref{empiric} presents the empirical
solutions yielded by the RNN\ for $d=2$, and manually \textquotedblleft
distilled\textquotedblright\ configurations. This includes the dimensionally
reduced model utilized in Sections \ref{sec:empirical_verification} \ and %
\ref{sec:invariance}, respectively, for the graphical and analytical proofs
of zero-error topological invariance, as well as the existence of a nonempty
open set of exact solutions. Sections \ref{sec:exact_solution_d3} and \ref%
{sec:exact_solution_d4} extend these exact results to $d=3$ and $d=4$.
Section \ref{sec:binary_multiplication_extension} presents the
neuro-symbolic extension to binary multiplication. In Section \ref{conj} we
argue that the results of this paper hint at a \textit{Universal} \textit{%
Topological Quantization Conjecture}, stating that for any discrete
algorithmic task, the continuous parameter space of an RNN inherently
harbors a topologically open, macroscopic subset of exact solutions
spontaneously reachable via optimized gradient descent. Finally, Section \ref%
{superAI} discusses the physics of amplification and the verticalization of
AI, and Section \ref{sec:conclusions} concludes the paper. In Appendices \ref%
{A} and \ref{B}, we compile a number of empirical network parameter sets
generated by the RNN via stochastic gradient descent in $d=2$, as well as
the manually distilled\ configurations obtained from them.

\section{Model Architecture and General Latent Dynamics}

\label{sec:model_architecture}

To establish a clear framework for mechanistic interpretability, we define a
minimal Recurrent Neural Network (RNN) designed to execute bit-serial binary
addition with carry.

\subsubsection*{A Structural Roadmap: Theory, Execution, and Training}

Before delving into the specific mathematical mechanics, it is crucial to
provide a structural roadmap of the framework. The task is decomposed into
three distinct conceptual layers, which will be addressed in the subsequent
sections:

\begin{enumerate}
\item The Theoretical Target (The Logical Ground-Truth): This represents
what we mathematically expect the system to achieve. It defines the strict
Boolean logic of positional binary addition---translating the abstract rules
of exact arithmetic into a deterministic, step-by-step sequence of target
sum bits and necessary carry transitions.

\item The Mechanical Execution (Operational Inference): This defines how the
network computes once its parameters are fixed. It encompasses the
continuous recurrent equations, the evolution of the latent state $h_{t}$,
the static weight matrices $W$ and bias vectors $b$, and the final
deterministic output rule.

\item The Training Phase (Geometric Optimization): This outlines how the
network discovers the correct weights $W$ and biases $b$ starting from a
completely unconfigured, random state. It details the optimization process
over the training dataset, the necessity of mapping raw logits into
continuous probabilities, and the introduction of the repulsive dead-zone
loss to safely mold the network's geometry without causing mathematical
degeneration.
\end{enumerate}

\subsubsection*{Sequence Representation and Dataset Formatting}

\label{sec:sequence_dataset}

To establish a transparent mapping between positional binary arithmetic and
temporal inputs, the model processes two binary sequences $%
A=(a_{1},a_{2},\dots ,a_{N})$ and $B=(b_{1},b_{2},\dots ,b_{N})$,
representing the positional bits of the two addends being summed. If the
original operands have different bit-lengths, the shorter sequence is
zero-padded to match the length of the longer one. Unlike standard human
mathematical notation, which reads binary numbers from left to right,
starting with the Most Significant Bit (MSB) down to the Least Significant
Bit (LSB), the RNN processes the inputs in reverse order. Specifically, the
bits are presented sequentially from LSB to MSB across discrete clock cycles 
$t=1,2,\cdots ,N$.

At discrete clock cycle $t=1$, the network receives the least significant
bits $a_{1},b_{1}\in \{0,1\}$ (corresponding to positional weight $2^{0}$),
followed at $t=2$ by $a_{2},b_{2}$ ($2^{1}$), and continuing sequentially up
to $t=N$. Furthermore, to accommodate potential carry overflow at the most
significant bit position without truncating the output, every binary
sequence is appended with a terminal zero-padding pair: $%
(a_{N+1},b_{N+1})=(0,0)$. This provides the recurrent state space with an
additional clock cycle to discharge any persistent internal carry state $%
c_{N}\in \{0,1\}$ and emit the final overflow sum bit $y_{N+1}$.

Each training instruction in the dataset is stored as a single formatted
text line:

\begin{center}
\texttt{input\_pairs : target\_sequence}
\end{center}

\noindent where \texttt{input\_pairs} is a comma-separated chronological
sequence of 2-bit vectors presented at each clock cycle $t$, and \texttt{%
target\_sequence} is the exact bit-serial target sequence expected from the
readout layer.

In the first stage of the investigation, the dataset used for training
exhaustively contained all 256 possible operations involving the sum of two
binary numbers with at most 4 bits ($16\times 16$ input pairs).
Subsequently, the dataset was reduced to the sums of binary numbers with at
most 3 and 2 bits ($8\times 8$ and $4\times 4$ input pairs, respectively).

\begin{table}[tb]
\centering%
\begin{tabular}{|c|c|c|c|c|c|}
\shortstack{Clock\\Cycle $t$ } & \shortstack{Input\\Pair $(a_{t},b_{t})$} & %
\shortstack{Prev.\\Carry ($c_{t-1}$)} & Sum & 
\shortstack{Output\\Bit
($y_{t}$)} & \shortstack{Next\\Carry ($c_{t}$)} \\ \hline
$t=1$ & \texttt{11} & $c_{0}=0$ & 10 & $y_{1}=0$ & $c_{1}=1$ \\ 
$t=2$ & \texttt{00} & $c_{1}=1$ & 01 & $y_{2}=1$ & $c_{2}=0$ \\ 
$t=3$ & \texttt{01} & $c_{2}=0$ & 01 & $y_{3}=1$ & $c_{3}=0$ \\ 
$t=4$ & \texttt{10} & $c_{3}=0$ & 01 & $y_{4}=1$ & $c_{4}=0$ \\ 
$t=5$ (Padding) & \texttt{00} & $c_{4}=0$ & 00 & $y_{5}=0$ & $c_{5}=0$%
\end{tabular}%
\caption{Step-by-step execution trace for the dataset example \texttt{%
11,00,01,10,00 : 01110} ($9+5=14$)}
\label{tab:1}
\end{table}

\paragraph{Concrete Dataset Example:}

Consider the representative dataset line:

\begin{center}
\texttt{11,00,01,10,00 : 01110}
\end{center}

This entry encodes the step-by-step addition of two 4-bit numbers ($N=4$,
padded to $N+1=5$ time steps):

\begin{enumerate}
\item Extracting the binary sequence $A$\textbf{:} Taking the first bit of
each pair yields the sequence $(1,0,0,1,0)$. Reversing from LSB-first to
standard MSB-first order gives the addend $1001$, which corresponds to the
decimal number $9$.

\item Extracting the binary sequence $B$: Taking the second bit of each pair
yields $(1,0,1,0,0)$. Reversing to standard order gives $101$, which
corresponds to the decimal number $5$.

\item Decoding the target sum: The target sequence \texttt{01110} represents
the expected output $y_{t}$ at each clock cycle $t$ ($y_{1}=0$, $y_{2}=1$, $%
y_{3}=1$, $y_{4}=1$, $y_{5}=0$). Reversing from LSB-first to standard order
yields the binary number $01110=1110$, which evaluates to the decimal number 
$14$.
\end{enumerate}

Step-by-step, the internal carry $c_{t}$ and sum bit $y_{t}$ evolve as shown
in Table \ref{tab:1}. This verifies the exact decimal arithmetic $9+5=14$.

\subsection{The Mechanical Execution (Operational Inference)}

In this subsection we explain how the network operates.

\subsubsection*{Input Representation and Latent State}

At each time step $t$, the neural network receives a two-dimensional binary
input vector $x_{t}=(x_{1,t},x_{2,t})^{T}\in \{0,1\}^{2}$, where its
components $x_{1,t}=a_{t}$ and $x_{2,t}=b_{t}$ correspond precisely to the $%
t $-th bits $a_{t}$ and $b_{t}$ of the two addends.

The internal memory of the network is encoded within a vector $h_{t}$
belonging to a continuous $d$-dimensional latent space $[-1,1]^{d}$, where $%
d\geq 2$ represents the latent dimension of the hidden state: 
\begin{equation}
h_{t}=\left( h_{1,t},h_{2,t},\cdots ,h_{d,t}\right) ^{T}\in \lbrack
-1,1]^{d}.
\end{equation}

\subsubsection*{Recurrent Update and Readout Equations}

The continuous state vector evolves dynamically according to the standard
non-linear recurrent equation%
\begin{equation}
h_{t}=\tanh \left( W_{ih}x_{t}+W_{hh}h_{t-1}+b_{h}\right)
\label{eq:general_hidden_update}
\end{equation}%
governed by the hyperbolic tangent activation function, applied
element-wise, which means $\tanh (v_{1},v_{2},\ldots )\equiv (\tanh
(v_{1}),\tanh (v_{2}),\ldots )$.

The trainable structural parameters consist of:

\begin{enumerate}
\item $W_{ih}\in \mathbb{R}^{d\times 2}$: the input-to-hidden weight matrix,
mapping external binary inputs into the latent space;

\item $W_{hh}\in \mathbb{R}^{d\times d}$: the hidden-to-hidden recurrent
matrix, encoding memory persistence and transition thresholds;

\item $b_{h}\in \mathbb{R}^{d}$: the hidden bias vector, anchoring the
resting state of the attractor.
\end{enumerate}

The internal state $h_{t}$ is subsequently mapped to output logits via a
linear readout layer: 
\begin{equation}
y_{t}=\left( y_{1,t}(h_{t}),y_{2,t}(h_{t})\right) ^{T}=W_{fc}h_{t}+b_{fc},
\label{eq:general_readout}
\end{equation}%
where $W_{fc}\in \mathbb{R}^{2\times d}$ is the fully-connected readout
matrix, and $b_{fc}\in \mathbb{R}^{2}$ is the output bias vector.

The predicted binary sum bit $\hat{y}_{t}\in \{0,1\}$ at cycle $t$ is
determined by the sign of the logit difference: 
\begin{equation}
\hat{y}_{t}=\frac{1}{2}\left[ \text{sign}(y_{2,t}-y_{1,t})+1\right] .
\label{eq:output}
\end{equation}%
The model returns a logical output bit $\hat{y}_{t}=1$ when $%
y_{1,t}(h_{t})<y_{2,t}(h_{t})$, and an output bit $\hat{y}_{t}=0$ when $%
y_{1,t}(h_{t})>y_{2,t}(h_{t})$.

\subsubsection*{The Theoretical Target (The Logical Ground-Truth)}

The latent vector $h_{t}$ serves as the complete continuous internal state
of the recurrent dynamical system at cycle $t$.

Letting $c_{t-1}$ denote the internal carry inherited from cycle $t-1$,
implicitly encoded within $h_{t-1}$ (as explained in detail below), the
continuous geometric state $h_{t}$ simultaneously:

\begin{enumerate}
\item Provides the spatial representation for the readout layer to decode
the predicted sum bit $\hat{y}_{t}\in \{0,1\}$, which is supposed to emulate
the target Boolean operation 
\begin{equation}
y_{t}^{\ast }=S_{t}+c_{t-1}\text{ mod 2}\in \{0,1\},  \label{targetlogic}
\end{equation}%
where $S_{t}=x_{1,t}+x_{2,t}\in \{0,1,2\}$ denotes the total input sum at
step $t$.

\item Serves as the persistent internal state that geometrically encodes the
updated carry 
\begin{equation}
c_{t}=\left\lfloor \frac{S_{t}+c_{t-1}}{2}\right\rfloor \in \{0,1\},
\label{upcarry}
\end{equation}%
propagating it forward across clock cycles to update the next latent state $%
h_{t+1}$.
\end{enumerate}

In turn, $h_{t+1}$ integrates the new input $S_{t+1}$ together with the
carry $c_{t}$ inherited from $h_{t}$ to decode the subsequent output bit $%
\hat{y}_{t+1}$ and generate the new carry $c_{t+1}$.

While the equations (\ref{targetlogic}) and (\ref{upcarry}) represent the
ideal Boolean logic of a perfect adder, the actual outputs generated by the
RNN are governed by its continuous state $h_{t}$ and readout equations (\ref%
{eq:output}).

Crucially, the carry $c_{t-1}$ is never explicitly provided by or to the
network; instead, it remains \textquotedblleft hidden\textquotedblright\ as
a geometric property within $h_{t-1}$ itself, which is precisely why the $h$
space is a \textit{latent }space. The idea is that, given a sufficiently
rich set of parameters in (\ref{eq:general_hidden_update}) and (\ref%
{eq:general_readout}), the continuous architecture can internally structure
the desired logic.

The RNN acts as a zero-error infinitely stable binary adder if the
continuous predictions generated by the neural network perfectly coincide
with the exact Boolean targets (\ref{targetlogic}) and (\ref{upcarry}) for
any arbitrary sequence length $t$.

A consequence of this structure is that the latent space dimension cannot be
lower than two ($d\geq 2$). Executing bit-serial addition with carry
requires maintaining at least two independent logical degrees of freedom:
the current output bit decision and the persistent carry state. A scalar
state space ($d=1$) lacks the geometric dimensionality needed to separate
the output hyperplanes from the carry hysteresis loop without state
collisions, establishing $d=2$ as the absolute minimal latent dimension for
a functional Recurrent Neural Network adder.

The formulation just given provides a general mathematical framework for the
entire study. It serves as the necessary basis to analyze empirical
solutions, search for spatial reductions (such as the 2D $\rightarrow $ 1D
scalar compression), and analytically prove infinite exactness in $d=2$
(Sections \ref{sec:empirical_verification} and \ref{sec:invariance}).
Furthermore, it enables the achievement of analogous results in higher
dimensions $d=3$ and $d=4$ (Sections \ref{sec:exact_solution_d3} and \ref%
{sec:exact_solution_d4}). Finally, it lays the groundwork to formally prove
the topological invariance and noise immunity of these embedded finite-state
machines across open subsets of the parameter space.

\section{The Training Phase (Geometric Optimization)}

\label{sec:dead_zone_loss}

In this section, we detail the geometric optimization process through which
the network discovers the structural parameters $W$ and $b$ starting from a
random, unconfigured state.

\subsection{Differentiable Output Probabilities and Training Logic}

The reliance of the deterministic decision rule (\ref{eq:output}) on the
discontinuous sign function renders it incompatible with gradient descent
optimization, as its derivative is zero almost everywhere \cite%
{Casey1996,Goodfellow2016,Trask2018}. To enable continuous training from an
unconfigured state with randomly initialized weights, the raw logit vector $%
y_{t}$ is mapped into a probability distribution.

Specifically, at each sequence step $t$, the network assigns the probability 
\begin{equation}
P(\hat{y}_{t}=k\hspace{0.02in}\hspace{0.01in}|\hspace{0.02in}\hspace{0.01in}%
x_{\leq t})=\frac{\mathrm{e}^{y_{k+1,t}(h_{t})}}{\mathrm{e}^{y_{1,t}(h_{t})}+%
\mathrm{e}^{y_{2,t}(h_{t})}}  \label{prediction}
\end{equation}%
to each binary output $k\in \{0,1\}$, via the standard softmax activation
applied to the linear readout $y_{t}$ of formula (\ref{eq:general_readout}) 
\cite{Goodfellow2016}. When the RNN is deployed for operational inference,
choosing the most probable class from this continuous distribution returns
the exact same logical output as the deterministic, sign-based decision rule
defined in Eq. (\ref{eq:output}).

Having defined $y_{t}^{\ast }$ as the ground-truth binary target at sequence
step $t$\ in (\ref{targetlogic}), $p_{t}=P(\hat{y}_{t}=y_{t}^{\ast }\mid
x_{\leq t})$ represents the network's estimated output probability for the
correct target class.

To optimize the structural parameters of the Recurrent Neural Network, we
must define the mechanical process of \textit{training}. Before training
begins, the network's internal weight matrices $W$ and bias vectors $b$ are
initialized with random values. Consequently, the initial predictions (\ref%
{prediction}) generated by the network are also random, and the
corresponding correct-class probabilities $p_{t}$ remain uninformative,
hovering near pure chance ($p_{t}\approx 0.5$). The goal of the optimization
process is to iteratively adjust the geometric parameters $W$ and $b$ until
the network's continuous predictions perfectly align with the exact
ground-truth outputs provided by the training dataset for the bit-serial
binary addition task.

This optimization is performed iteratively by processing the dataset in
mini-batches. Let $B$ be the batch size, defined as the number of
independent data samples (i.e., $B$ distinct arithmetic operations of the
dataset) evaluated simultaneously in a single computational step. We recall
that $L=N+1$ is the temporal sequence length in clock cycles (corresponding
to the bit-length $N$ of the operands plus the final carry discharge cycle).
For a given batch of size $B$, we compare the network's continuous
predictions $p_{t}$ against the ground-truth targets by computing the total
loss function $\mathcal{L}_{\text{tot}}$.

This function is defined as the mean scalar evaluation of the individual
penalty $\mathcal{L}(p_{n,t})$ (defined below) across all $B$ data samples
within the batch\ (labelled by the subscript $n$) and their respective $%
L_{n} $ clock cycles (labelled by $t$): 
\begin{equation}
\mathcal{L}_{\text{tot}}=\frac{1}{\sum_{n=1}^{B}L_{n}}\sum_{n=1}^{B}%
\sum_{t=1}^{L_{n}}\mathcal{L}(p_{n,t}).  \label{Lto}
\end{equation}%
The individual penalty $\mathcal{L}(p_{n,t})$ is calculated for a single
predicted bit at a specific clock cycle $t$ within a specific data sample $n$%
.

By computing the gradient of the total loss $\mathcal{L}_{\text{tot}}$ with
respect to each network parameter $w$ (representing both weights $W$ and
biases $b$), the optimization algorithm dictates how the geometry is
adjusted.

Following the steepest descent principle, the parameter update rule $%
w\rightarrow w+\Delta w$ at each computational step is determined by: 
\begin{equation}
\Delta w=-\eta \frac{\partial \mathcal{L}_{\text{tot}}}{\partial w},
\label{c1}
\end{equation}%
where $\eta >0$ is the \textit{learning rate}---an empirical scalar
hyperparameter that specifies the step size at each iteration and can either
be kept fixed or dynamically modulated during training.

This update mechanism guarantees that the variation of the total loss is
non-positive to the lowest order in Taylor expansion: 
\begin{equation}
\Delta \mathcal{L}_{\text{tot}}\simeq \sum_{w}\frac{\partial \mathcal{L}_{%
\text{tot}}}{\partial w}\Delta w=-\eta \sum_{w}\left( \frac{\partial 
\mathcal{L}_{\text{tot}}}{\partial w}\right) ^{2}\leq 0.  \label{c2}
\end{equation}%
However, because this inequality strictly holds only for infinitesimal
parameter updates, practical optimization with a finite learning rate $\eta $
inevitably introduces higher-order effects. As a result, the actual loss $%
\mathcal{L}_{\text{tot}}$ can temporarily increase. Far from being a
computational flaw, this transient non-monotonic behavior is structurally
necessary, because it provides the dynamical system with the energy required
to escape approximate local minima and successfully reach the globally
stable geometric attractors.

By iteratively applying these updates across successive batches, the
continuous dynamical system learns to minimize the penalty, progressively
molding its geometry to encode the exact logical algorithm.

At the beginning of the optimization process, the weights are random,
causing the logits to be tightly clustered around zero; consequently, the
initial output probabilities lack any statistical confidence ($p_{t}\approx
0.5$). The goal of the optimization algorithm is to adjust the geometric
parameters to confidently drive $p_{t}$ close enough to $1$ for all steps in
the sequence.

The most common approach to train neural networks for predicting discrete
targets (such as binary outputs) is to minimize the Cross-Entropy loss. For
a single sequence step $t$, the standard penalty is simply the negative
logarithm of the correct class probability:%
\begin{equation}
\mathcal{L}_{\text{CE}}(p_{t})=-\log (p_{t}).  \label{CEloss}
\end{equation}%
The logic behind this common loss function is straightforward: as the
network's confidence in the correct answer increases ($p_{t}\rightarrow 1$),
the penalty approaches zero. Conversely, if the network is highly uncertain
or confident in the wrong answer ($p_{t}\rightarrow 0$), the penalty
approaches infinity, generating a massive gradient to correct the weights
according to the update rule (\ref{c1}).

However, applying the standard Cross-Entropy loss (\ref{CEloss}) to exact
algorithmic tasks is fundamentally flawed for two distinct reasons.

Consider the situation where the network already predicts the correct bit
with 95\% confidence ($p_{t}=0.95$): the logical output is already perfect.
In discrete binary logic, the task is successfully executed as long as the
probability of the correct answer safely exceeds that of the incorrect one.
Yet, because the derivative of $-\log (p_{t})$ never evaluates to zero, it
still exerts a force on the parameters to push this probability indefinitely
closer to $1$. The optimization algorithm keeps modifying the parameters in
a perpetual attempt to reach exactly 100\%. This numerically decreases the
loss, creating a mathematical illusion of progress while the actual logical
accuracy remains identically unchanged. In rigid algorithmic tasks, this
perpetual gradient push keeps the latent space in a state of constant flux,
preventing the dynamical system from settling into a stable geometric
attractor.

Second, forcing the probability asymptotically to 100\% causes the
continuous phase space to geometrically degenerate. This phenomenon is
analogous to severe polynomial overfitting. To extract extreme precision on
the specific training examples, the gradient drives the parameter weights
toward large values, violently saturating the hyperbolic tangent
activations. This creates chaotic, fragile decision boundaries. In this
situation, the network may perfectly memorize the training set, but any
microscopic floating-point truncation noise introduced during extended
continuous execution will shift the state vector slightly. Because the
overfitted geometric boundaries are so brittle, these nearby perturbed
points derail the continuous trajectory, leading to catastrophic logical
failures.

\subsection{The Repulsive Dead-Zone Loss Function}

Standard gradient descent optimization of Recurrent Neural Networks applied
to rigid logical tasks frequently stagnates in approximate local minima,
yielding continuous trajectories that fail to achieve zero error across
arbitrary sequence lengths~\cite{Bengio1994, Hochreiter1997,Kaiser2015}. To
overcome the difficulties described so far and force the continuous latent
dynamics to quantize into discrete symbolic logic, in this work we introduce
a custom saturated cost function, which we term the \textit{repulsive
dead-zone loss. }The penalty is specifically designed to safely mold the
network's continuous geometry without causing mathematical degeneration.

The repulsive dead-zone loss function $\mathcal{L}_{\text{RDZ}}(p_{t})$ must
achieve two conflicting goals: it must stop adjusting the weights once a
safe logical margin is reached (to allow stabilization), and it must
actively repel the parameters to prevent degenerate geometric saturation
(overfitting), thereby forcing the continuous phase space to quantize into
infinitely stable, error-absorbing basins of attraction that constrain the
system's trajectories within logical zones separated by a finite gap and
mapped into their interiors with a positive safety margin.

We define the penalty as the piecewise smooth function 
\begin{equation}
\mathcal{L}_{\text{RDZ}}(p_{t})=%
\begin{cases}
-\log (p_{t}), & \text{if }p_{t}<\theta _{1}, \\[6pt] 
0, & \text{if }\theta _{1}\leq p_{t}\leq \theta _{2}, \\[6pt] 
-\log (1-p_{t}+\epsilon ), & \text{if }p_{t}>\theta _{2},%
\end{cases}
\label{eq:dead_zone_loss}
\end{equation}%
where $\theta _{1}=0.90$ defines the lower confidence threshold, $\theta
_{2}=0.98$ specifies the upper saturation threshold, and $\epsilon =10^{-7}$
ensures numerical stability.

Unlike standard Cross-Entropy loss, which maintains non-zero gradients even
for well-classified samples, $\mathcal{L}_{\text{RDZ}}$ creates a dual
optimization regime:

\begin{enumerate}
\item Active Training Region ($p_{t}<0.90$)\textbf{:} Standard logarithmic
gradients drive low-confidence predictions toward the correct logical
regions.

\item Neutral Dead Zone ($0.90\leq p_{t}\leq 0.98$)\textbf{:} Gradient
forces vanish, allowing parameter updates for well-performing samples to
coast without unnecessary perturbation.

\item Saturation Driving Region ($p_{t}>0.98$)\textbf{:} The inverted
log-penalty $-\log (1-p_{t}+\epsilon )$ exerts a targeted repulsive push
that prevents extreme saturation and artificial overconfidence, acting as an
implicit regularizer that helps mitigate overfitting. This force is
calibrated just enough to lock the continuous latent dynamics into invariant
basins of attraction embedding compact logical zones separated by a finite
gap and mapped into their interiors, shielding state transitions against
numerical noise.
\end{enumerate}

The total loss (\ref{Lto}) utilized in this paper over a batch of size $B$
composed of sequences of varying lengths $L_{n}$ is computed as the average 
\begin{equation}
\mathcal{L}_{\text{tot}}=\frac{1}{\sum_{n=1}^{B}L_{n}}\sum_{n=1}^{B}%
\sum_{t=1}^{L_{n}}\mathcal{L}_{\text{RDZ}}(p_{n,t})
\end{equation}%
of $\mathcal{L}_{\text{RDZ}}(p_{n,t})$ across all data samples within the
batch (indexed by $n$) and their respective $L_{n}$ clock cycles (indexed by 
$t$).

\subsection{Relation to Grokking, Regularization and Algorithmic
Generalization}

The multi-stage optimization dynamics observed here---characterized by an
extended period of low-error plateauing followed by an abrupt transition to
zero-error generalization---exhibits parallels with the phenomenon known as
grokking, first formalized by Power et al. \cite{power} in the context of
modular arithmetic and small algorithmic datasets, subsequently explored in
recurrent neural networks by Murray \cite{murray}, and contextualized within
systematic regularization frameworks by Manir and Rupa \cite{grokking}.
While standard deep learning intuition expects generalization to unseen
sequences beyond the training dataset to happen alongside training
convergence, algorithmic tasks often exhibit a decoupled regime where the
network interpolates the learned data long before discovering the underlying
structural rule. Furthermore, standard practices in deep learning frequently
address aspects of training pathology via conventional toolsets---such as
mitigating probability saturation or employing margin-based penalties.
Instead, our framework resolves a fundamentally different regime unique to
sequential recurrent dynamics.

Specifically, three core distinctions separate standard approaches from our
formulation:

1) Regarding the optimization objective, phenomena like overconfidence and
asymptotic probability saturation are widely recognized in static
classification \cite{Goodfellow2016}, such as image recognition, spam
filtering, sentiment analysis, or medical diagnostics. Nevertheless, their
destructive impact in RNNs executing sequential algorithms goes far beyond
statistical miscalibration. In static feed-forward models, standard
cross-entropy pushes logits toward infinity without engaging a temporal
sequence; in algorithmic RNNs, however, continuous backpropagation of these
unconstrained, infinite-magnitude gradients across temporal clock cycles
generates a chaotic drift within the continuous latent space, destroying the
long-term memory persistence required for the carry state. Standard
cross-entropy fundamentally fails because it treats discrete time steps as
independent classification tasks, entirely ignoring the geometric coherence
required to maintain stable dynamical attractors over unbounded horizons.

2) While margin-based penalties (such as hinge loss in support vector
machines) and probability-smoothing techniques have long been utilized in
static classification to penalize excessive confidence or regularize
statistical generalization, the repulsive dead-zone loss $\mathcal{L}_{\text{%
RDZ}}$ acts as a geometric sculptor for dynamical systems rather than a mere
external regularizer. By establishing a zero-gradient neutral zone coupled
with an asymptotic repulsive barrier, it actively forces the continuous
phase space of a recurrent network to undergo a global topological
bifurcation---transforming the empirical, unpredictable nature of standard
grokking into a controlled, mathematically certified topological phase
transition. This carves out robust attractors capable of absorbing
floating-point truncation noise across infinite temporal trajectories.

3) While the literature predominantly investigates static architectures
(such as feedforward MLPs or Transformers processing atemporal lookup
tables), our work applies this phase-transition dynamics to continuous
recurrent systems, demonstrating how a minimal RNN builds, maintains, and
protects a persistent long-term memory state (such as the sequential carry
bit) via low-norm geometric attractors to execute time-dependent algorithms.

\section{Loss Dynamics and Phase Transition in \texorpdfstring{$d=2$}{d=2}}

\label{empiric}

In this section, we switch from the general theoretical framework of the
repulsive dead-zone loss to its concrete empirical application within the
minimal two-dimensional latent space ($d=2$).

We first detail the training dynamics, highlighting the topological phase
transition to Boolean logic. Subsequently, following numerical convergence,
we extract and manually refine the learned parameters to derive a
\textquotedblleft distilled model\textquotedblright , which serves as the
foundation for the rigorous graphical and analytical proofs of zero-error
infinite stability presented in Sections \ref{sec:empirical_verification} \
and \ref{sec:invariance}.

The network parameters are initialized randomly with small Gaussian weights (%
$W\sim \mathcal{N}(0,0.01)$, where $\mathcal{N}(\mu ,\sigma ^{2})$ denotes
the normal distribution with mean $\mu $ and variance $\sigma ^{2}$) and
zero biases $b$. Training is conducted using the Adaptive Moment Estimation
(Adam) optimizer on the binary addition dataset.

We begin with the complete 256-instance dataset for 4-bit operands ($%
16\times 16$ input pairs) with a mini-batch size of $B=64$. The training is
later repeated on reduced domains of 64 instances (3-bit operands, $B=64$)
and 16 instances (2-bit operands, $B=16$).

To maintain stable gradient flow while enabling stochastic exploration
across flat landscape regions, we employ a multi-cycle Cosine Annealing
learning rate schedule starting from an initial learning rate $\eta
_{0}=0.005$. Optimization is governed by the custom repulsive dead-zone loss
function $\mathcal{L}_{\text{RDZ}}$ defined in Eq. (\ref{eq:dead_zone_loss}%
). 
\begin{table}[t]
\centering{\small 
\begin{tabular}{lcccccc}
\toprule Epoch & 1 & 440 & 834 & 6455 & 8221 & 8711 \\ 
Loss $\mathcal{L}$ & 0.692694 & 0.103498 & 0.067283 & 0.017953 & 0.004266 & 
0.010087 \\ 
\midrule Epoch & 8712 & 8713 & 8714 & 8715 & 8716 & 8717 \\ 
Loss $\mathcal{L}$ & 0.003123 & 0.004248 & 0.000628 & 0.004453 & 0.001075 & 
0.000000 \\ 
\bottomrule &  &  &  &  &  & 
\end{tabular}
}
\caption{Empirical loss collapse observed during training with a
256-instance dataset. Around Epoch 8716, the loss abruptly drops to zero,
marking the completion of the topological phase transition to Boolean logic}
\label{tab:loss_collapse}
\end{table}

\begin{figure}[t]
\centering\includegraphics[width=1\linewidth]{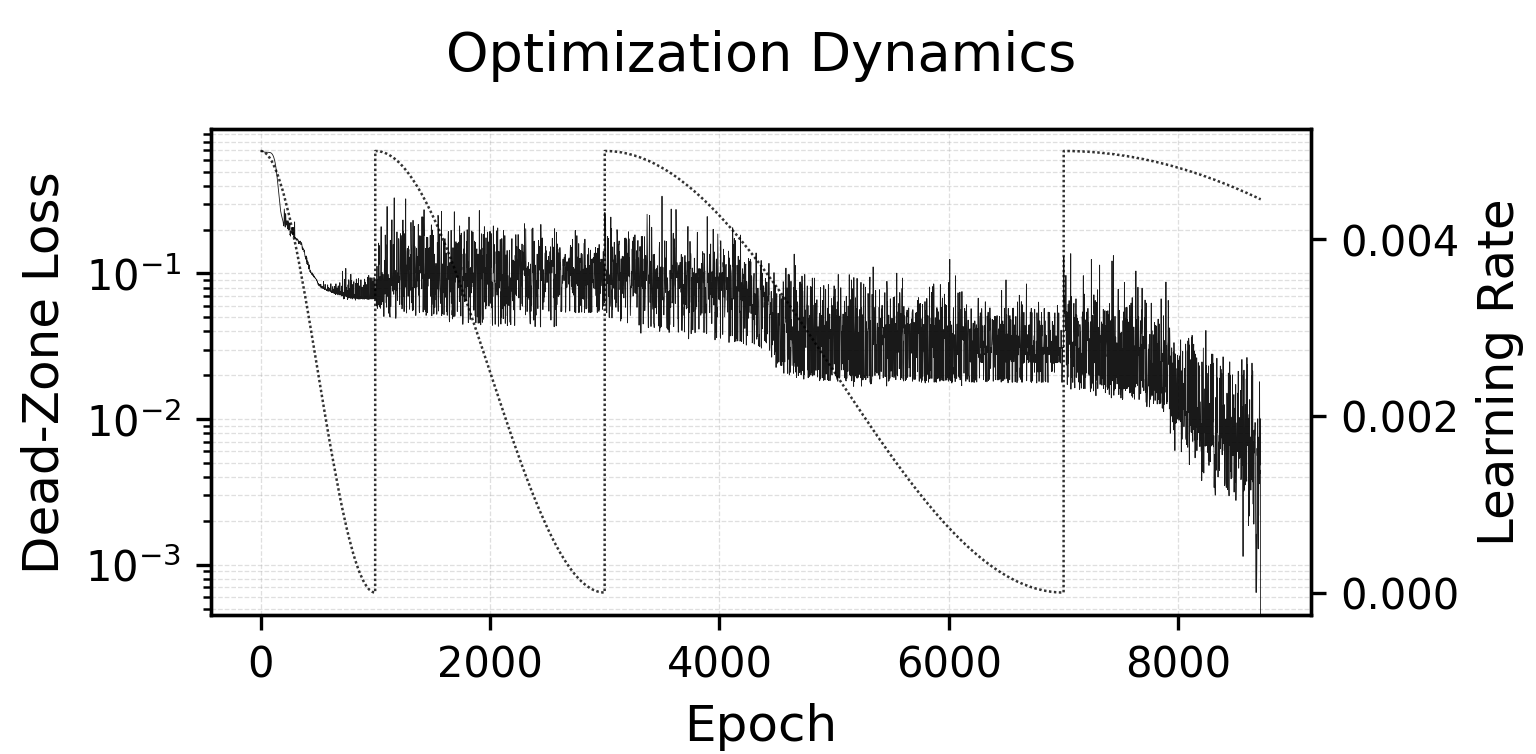}
\caption{Empirical training dynamics under the repulsive dead-zone loss $%
\mathcal{L}_{\text{RDZ}}$. The plot illustrates the multi-cycle Cosine
Annealing schedule (dotted line) and the corresponding log-scale loss
trajectory (solid line), highlighting the abrupt topological phase
transition to zero-error saturation around Epoch 8717}
\label{empir}
\end{figure}

As captured in our empirical training logs and shown in Table~\ref%
{tab:loss_collapse} and Fig. \ref{empir}, the optimization trajectory on the
complete dataset of 256 instances, corresponding to all possible additions
of binary numbers up to 4 bits, exhibits a multi-stage dynamic characterized
by periodic learning-rate restarts and a sudden topological phase
transition, unfolding as follows:

\begin{enumerate}
\item High-Error Initial Descent (Epochs 0--600): The loss drops rapidly
from $\mathcal{L}_{\text{tot}}\approx 0.69$ at the initial unconfigured
state down to $\mathcal{L}_{\text{tot}}\approx 0.08$ as gradients derived
from the logarithmic penalty drive low-confidence predictions toward the
correct target basins in the continuous latent space.

\item Multi-Cycle Dead-Zone Exploration (Epochs 600--8500): Under successive
Cosine Annealing cycles (with periodic resets at learning rate $0.005$),
predictions falling within $[0.90,0.98]$ enter the neutral dead-zone. The
loss oscillates on a low-error plateau $\mathcal{L}_{\text{tot}}\in \lbrack
0.015,0.08]$, preventing gradient chatter and allowing parameter weights to
stabilize without unnecessary perturbation.

\item Topological Phase Transition and Collapse to Boolean Logic (Epochs
8500--8717): As predictions approach the upper boundary ($p_{t}>0.98$), the
inverted log-penalty acts as a repulsive barrier, actively preventing the
parameters from exploding and the tanh activations from degenerating into
extreme saturation. Confined by this topological boundary, the optimization
landscape undergoes a sharp autocatalytic collapse within just one epoch
(specifically, between Epoch 8716 and Epoch 8717), reaching $\mathcal{L}_{%
\text{tot}}\simeq 0$. As all sequence trajectories are safely locked within
the zero-penalty dead zone, the continuous latent space abruptly fractures
into globally stable, invariant basins that trap the system into compact
logical zones separated by a finite gap and mapped into their interiors by
the dynamics.
\end{enumerate}

\subsection{Thermodynamics of the Topological Phase Transition}

\label{thermo}

Under the influence of the repulsive dead-zone loss (\ref{eq:dead_zone_loss}%
) the system exhibits a characteristic \textit{autocatalytic collapse},
signaling that the continuous latent space $(h_{1},h_{2})\in \lbrack
-1,1]^{2}$ quantizes into a globally stable discrete finite-state machine.
Rather than stagnating in approximate local minima, after thousands of
epochs navigating non-convex landscapes with fluctuating loss values, the
loss drops vertically to numerical zero ($\mathcal{L}<10^{-6}$): the network
suddenly discovers a critical geometric bottleneck and achieves zero-error
execution across arbitrary sequence lengths.

We observed analogous autocatalytic collapses with datasets of sizes 64 and
16 (corresponding to all possible additions of binary numbers with up to 3
or 2 bits), for which the mini-batch sizes $B$ were correspondingly set to
64 and 16.

Training on the 64-addition dataset revealed no significant dynamical
differences compared to the thermodynamic trajectory described above for the
256-addition dataset. The topological collapse typically occurs within a few
minutes on standard GPU hardware.

With the dataset of 16 additions, conversely, the extreme sparsity of the
constraints required a structural adaptation of the optimization strategy to
bypass flat, uninformative regions of the parameter space. We employed a
hybrid optimization engine, alternating standard gradient descent with
surgical, single-parameter backtracking to actively suppress the Adam
optimizer's inertia. By modulating the learning-rate variation through a
multi-cycle schedule and extending the exploration indefinitely, the system
successfully navigated the geometric bottleneck. The resulting empirical
loss trajectory, detailed in Table \ref{3}, documents the sudden transition:
after fluctuating in a low-error plateau for hundreds of thousands of
epochs, the continuous dynamics abruptly quantized into an exact zero-error
state at epoch 302,834. 
\begin{table}[t]
\centering{\small {\ 
\begin{tabular}{cccccc}
\toprule Epoch & 302470 & 302510 & 302570 & 302610 & 302620 \\ 
Loss $\mathcal{L}$ & 0.005447 & 0.017109 & 0.007882 & 0.007550 & 0.101071 \\ 
Time (s) & 1897.8 & 1898.0 & 1898.5 & 1898.7 & 1898.8 \\ 
\midrule Epoch & 302760 & 302820 & 302830 & 302833 & 302834 \\ 
Loss $\mathcal{L}$ & 0.002406 & 0.002398 & 0.002398 & 0.002398 & 0.000000 \\ 
Time (s) & 1899.6 & 1900.0 & 1900.1 & 1900.1 & 1900.1 \\ 
\bottomrule &  &  &  &  & 
\end{tabular}
} }
\caption{Empirical loss collapse observed during hybrid optimization with
the highly sparse size 16 dataset. The abrupt drop to absolute zero marks
the topological phase transition}
\label{3}
\end{table}

An empirical zero-error state on finite training instances does not
automatically guarantee infinite-horizon stability or immunity to numerical
drift. To verify whether this loss collapse represents a true algorithmic
crystallization rather than a transient memorization artifact, the
continuous RNN predictions were strictly evaluated against the exact Boolean
logic of binary addition across extended sequence lengths.

Various network configurations are collected in Appendix \ref{A}, alongside
details of the extensive stress tests performed on them: all possible sums
of binary numbers up to and including 14 bits, long random sequences, and
adversarial edge cases. Across all evaluated sequences, no computational
failures were ever detected, confirming that zero-error infinite stability
has been reached.

After numerical convergence, we selected the checkpoints exhibiting the
clearest weight symmetries. By applying small manual refinements to their
weights, we derived \textquotedblleft distilled models\textquotedblright
---simplified configurations that successfully pass the same stress tests as
the original networks, and therefore serve as equally strong candidates for
zero-error infinite stability. The three primary distilled models are
detailed in Appendix \ref{B}.

In particular, the \textquotedblleft Distilled Model B\textquotedblright ,
obtained by training the network on the 256-addition dataset, is
characterized by the following parameters: 
\begin{eqnarray}
W_{ih} &=&%
\begin{pmatrix}
0.25 & \text{ }0.25 \\ 
-2.25 & \text{ }-2.25%
\end{pmatrix}%
,\qquad W_{hh}=%
\begin{pmatrix}
-0.4 & \text{ }-0.25 \\ 
3.5 & \text{ }2.1875%
\end{pmatrix}%
,\qquad W_{fc}=%
\begin{pmatrix}
0.0 & \text{ }0.0 \\ 
4.0 & \text{ }1.0%
\end{pmatrix}%
,  \notag \\
b_{h} &=&%
\begin{pmatrix}
-0.7 & \text{ }3.5%
\end{pmatrix}%
^{T},\qquad b_{fc}=%
\begin{pmatrix}
0.0 & \text{ }1.0%
\end{pmatrix}%
^{T}.  \label{solution}
\end{eqnarray}%
This configuration exhibits an exact 1D scalar dynamic ($%
Z_{t}=-0.4h_{1,t}-0.25h_{2,t}$), detailed in Section \ref%
{sec:empirical_verification}. A\ similar dimensional reduction of the latent
space is observed in Distilled Model C, which was derived by training the
network on the 16-addition dataset.

In the next two sections we prove zero-error infinite stability of the model
(\ref{solution}), both graphically (Section \ref{sec:empirical_verification}%
) and analytically (Section \ref{sec:invariance}). We also demonstrate that
configuration (\ref{solution}) is not an isolated zero-error stable
arrangement, but rather belongs to an open set of such solutions. This
topological property is precisely what enables the RNN, once optimized via
the repulsive dead-zone loss function (\ref{eq:dead_zone_loss}), to reliably
converge to one of them.

Appendix \ref{B} contains the graphical proof of the same results for
Distilled Model C.

The collapse we observed across all evaluated cases exhibits the classic
thermodynamic signature of a \textit{topological phase transition}, a
phenomenon theoretically established in the context of finite-state automata
emulation within recurrent networks \cite{Casey1996}. However, while early
classical proofs relied on idealizing, physically inapplicable
assumptions---such as discontinuous step activations or infinite-precision
irrational weights---our framework demonstrates that common,
finite-precision continuous networks can autonomously self-organize into
exact, zero-error finite-state machines under standard \texttt{float32}
precision. This bridges the historical gap~\cite{Giles1992, Siegelmann1995}
between the abstract mathematical potential of RNNs and their real-world
trainability, proving that exact logical states are physically accessible
via optimized gradient descent.

While our terminology conceptually borrows from condensed matter physics,
within the strict framework of dynamical systems theory, the transition is 
\textit{topological} because it fundamentally reshapes the underlying
qualitative structure of the continuous latent phase space: rather than
undergoing mere incremental numerical adjustments, the system experiences a
global bifurcation. The smooth, connected $h$-manifold fractures into
discrete invariant basins (called $I_{0}$ and $I_{1}$ in the next sections).
Partitioned by the linear readout boundary, these basins identify four
compact logical quadrants $A_{ij}$, $i,j\in \{0,1\}$, which are separated by
finite gaps and mapped into their interiors by the dynamics. This locks the
attractor surface into a globally stable configuration of the finite-state
machine where all sequence evaluations simultaneously fall inside the
zero-penalty dead zone. Because this geometric separation shields the
internal states against perturbations and numerical noise, the continuous
latent dynamics quantize into an exact, error-tolerant discrete automaton.

\section{Graphical Proof of Zero-Error Topological Invariance in 
\texorpdfstring{$d=2$}{d=2}}

\label{sec:empirical_verification}

Solution (\ref{solution}) gives the following evolution and output logit:%
\begin{eqnarray}
h_{1,t} &=&\tanh \left( \frac{1}{4}S_{t}+Z_{t-1}-\frac{7}{10}\right) ,\qquad
h_{2,t}=\tanh \left( -\frac{9}{4}S_{t}-\frac{35}{4}Z_{t-1}+\frac{7}{2}%
\right) ,  \notag \\
y_{t} &=&4h_{1,t}+h_{2,t}+1,\qquad S_{t}\equiv x_{1,t}+x_{2,t},\qquad
Z_{t-1}\equiv -\frac{2}{5}h_{1,t-1}-\frac{1}{4}h_{2,t-1},  \label{system}
\end{eqnarray}%
while the output bit is 
\begin{equation}
\hat{y}_{t}=\frac{1}{2}\left( \text{sign}(y_{t})+1\right)  \label{outputbit}
\end{equation}%
by formula (\ref{eq:output}). Due to the rank-1 structure assumed by the
recurrent weight matrix $W_{hh}$, the two-dimensional latent space collapses
onto the scalar compression variable $Z_{t}$.

\begin{figure}[tb]
\centering\includegraphics[width=16truecm]{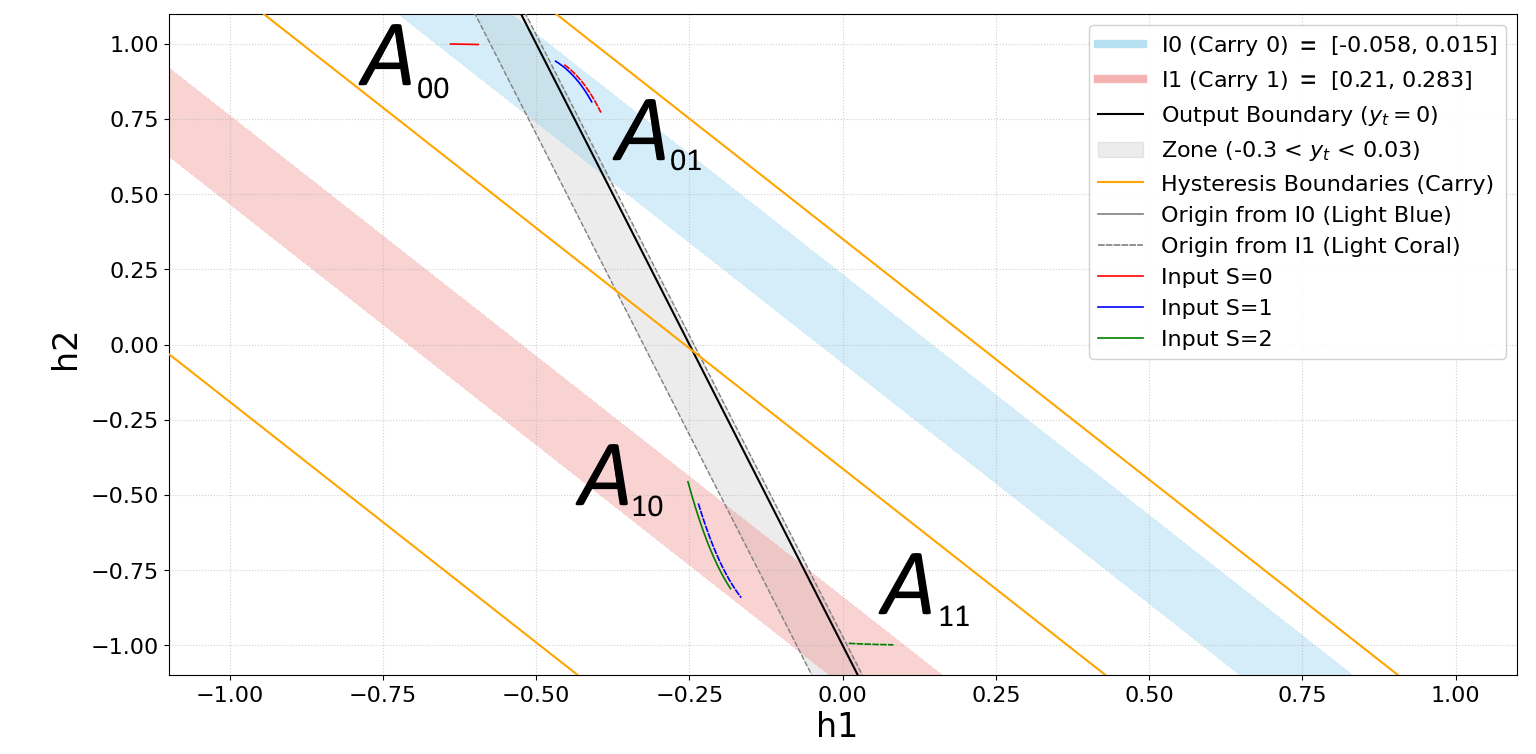}
\caption{Global phase space mapping of the invariant basins $I_{0}$
(skyblue) and $I_{1}$ (lightcoral) in $d=2$. The output boundary (black
solid line) and carry hysteresis boundaries (orange solid lines) partition
the space into multiple regions, four of which safely enclose the logical
target zones, denoted as $A_{ij}$, where $i$ is the next carry and $j$ is
the output bit. Trajectory arcs demonstrate how states originating from $%
I_{0}$ and $I_{1}$ under total input sum $S\in \{0,1,2\}$ correctly map
into\ the interiors of their respective $A_{ij}$ areas}
\label{graph1}
\end{figure}

\begin{figure}[tb]
\centering\includegraphics[width=16truecm]{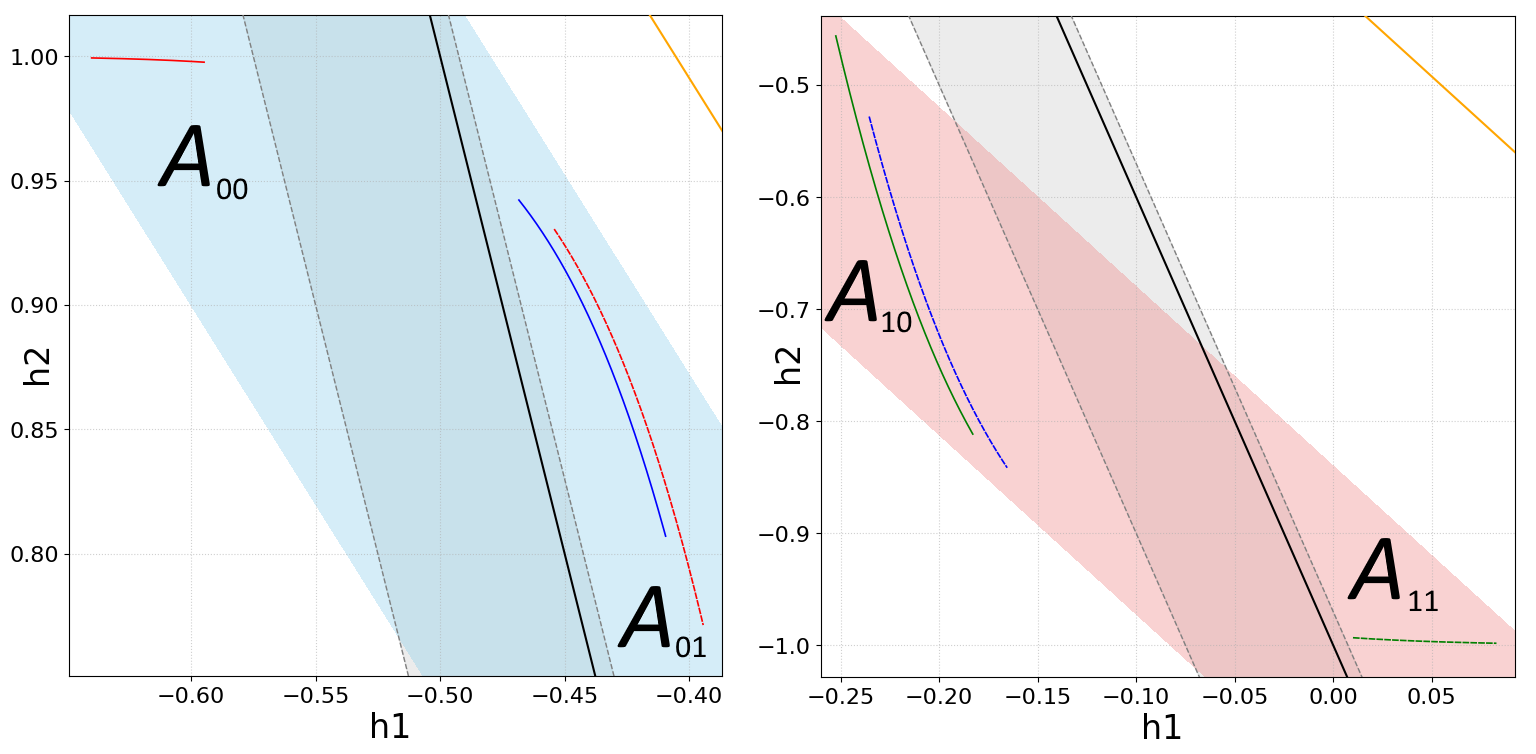}
\caption{Magnification of the key target areas. Mapped trajectories for
inputs $S_{t}=0$ (red), $S_{t}=1$ (blue), and $S_{t}=2$ (green) remain
strictly confined within the interiors of their expected arrival zones $%
A_{ij}$}
\label{graph2}
\end{figure}

With reference to Figs. \ref{graph1} and \ref{graph2}, the phase space $%
(h_{1},h_{2})\in \lbrack -1,1]^{2}$ is geometrically structured by two
distinct sets of boundaries. Defining 
\begin{equation}
Z(h)=-\frac{2}{5}h_{1}-\frac{1}{4}h_{2},\qquad Y(h)=4h_{1}+h_{2}+1,
\label{ZY}
\end{equation}%
we have:

1. \textit{Readout/output decision boundary}. This is the straight black
line $Y(h)=0$, i.e., 
\begin{equation}
4h_{1}+h_{2}+1=0.  \label{output}
\end{equation}%
It separates the region decoding to output bit $0$ ($Y<0$) from the one
decoding to output bit $1$ ($Y>0$).

2. \textit{Carry hysteresis boundaries}. These are the orange lines,
determined by the three critical roots $Z_{1}\approx -0.088$, $Z_{2}\approx
0.103$ and $Z_{3}\approx 0.448$ of the equation 
\begin{equation}
4\tanh \left( Z-\frac{9}{20}\right) +\tanh \left( \frac{5}{4}-\frac{35}{4}%
Z\right) +1=0,  \label{carry}
\end{equation}%
which delimits the transition thresholds for long-term carry memory.

The logical rationale underlying these two sets of boundaries is as follows.

We know that the output is encoded by a function $y_{t}(h_{t})$ of the
evolved latent state $h_{t}$, which is in turn a function $%
h_{t}(S_{t},h_{t-1})$ of the input $S_{t}$ and the previous latent state $%
h_{t-1}$. Given a generic point $h_{t-1}\in \lbrack -1,1]^{2}$, we need to
determine whether it is mapped into a point $h_{t}$ with output bit $\hat{y}%
_{t}$ equal to 0 or 1. This lays out two regions separated by the $%
y_{t}=Y(h_{t})=0$ line (\ref{output}): the region with output 0 ($y_{t}<0$)
is the one to the left of the black line in Figs. \ref{graph1} and \ref%
{graph2}, while the region with output 1 ($y_{t}>0$) the one to the right.

Definition 2 is explained by the logic of the neutral input $S_{t}=1$ and
memory testing. We must consider that the carry is a \textit{hidden}
function $c(h)$ over the latent space $h\in \lbrack -1,1]^{2}$. The RNN
autonomously uncovers it during training, as explained in previous sections.
Here we must make $c(h)$ mathematically explicit to provide a rigorous proof
of zero-error infinite stability.

The binary arithmetic (formulas (\ref{targetlogic}) and (\ref{upcarry}))
states that at $S_{t}=1$ the new carry is equal to the previous one, $%
c_{t}=c_{t-1}$, and the output $\hat{y}_{t}$ coincides with the flipped
previous carry. Hence, for every point $h_{t-1}\in \lbrack -1,1]^{2}$ in the
square, we define the candidate carry function%
\begin{equation}
c(h_{t-1})=\frac{1}{2}\left[ 1-\text{sign}(y_{t}(h_{t}(1,h_{t-1})))\right] ,
\label{carrycandidate}
\end{equation}%
obtained by flipping (\ref{outputbit}) at $S_{t}=1$, viewing the output
function $y_{t}$ as a function of $S_{t}$ and $h_{t-1}$ through $h_{t}$. The
decision boundary of $c(h)$ is given by $y_{t}(h_{t}(1,h))=0$, which
corresponds to Eq. (\ref{carry}). The orange lines in the figures separate
the regions with different carries, 0 or 1.

The proof that the proposed definition of carry indeed works goes along with
the proof that the system as a whole does what the binary arithmetic
prescribes. Moreover, although the definition (\ref{carrycandidate}) applies
to \textit{any} point $h_{t-1}\in \lbrack -1,1]^{2}$, it does not represent
the carry on the square as a whole, but only in the infinitely stable areas
that we are going to identify in a moment.

It is not true, for example, that at $S_{t}=1$ any point $h_{t-1}\in \lbrack
-1,1]^{2}$ is mapped to a point $h_{t}(1,h_{t-1})$ with the same carry. The
equality $c(h_{t})=c(h_{t-1})$ only holds in selected areas. In a continuous
dynamical system, embedded in a real space and governed by weight matrices
and non-linear hyperbolic tangents, the existence of a hidden function $c(h)$
that remains unaltered after a geometric transformation is the exception,
not the rule.

It is also worth noting that our choice (\ref{carrycandidate}) to define the
carry using the (inverted) output at $S_{t}=1$ is not unique. We could have
equivalently defined it using the output at $S_{t}=0$ (without inversion). A
direct correspondence also occurs for $S_{t}=2$. Probing the network with
any of these options leads to the same results, confirming that the hidden
carry is a robust physical property of the system, independent of the
specific way we choose to reveal it. 

At this point, we isolate the following memory basins along the scalar axis $%
Z(h)$: 
\begin{eqnarray}
Z &\in &I_{0}=[-0.058,0.015]\quad \text{(Inactive Carry / Carry 0),}  \notag
\\
Z &\in &I_{1}=[0.21,0.283]\quad \text{(Active Carry / Carry 1),}
\label{basins}
\end{eqnarray}%
alongside the following output basins for the readout logit $Y(h)$: 
\begin{equation}
Y\in J_{0}=[-4,-0.3]\quad \text{(Binary Output 0), \qquad }Y\in
J_{1}=[0.03,6]\quad \text{(Binary Output 1).}  \label{outbas}
\end{equation}%
The specific numerical values defining the intervals of (\ref{basins}) and (%
\ref{outbas}) are selected primarily to maximize the clarity of the
graphical proof. These boundaries are not strictly rigid. Rather, they are
intentionally conservative, establishing internal geometric margins that
safely avoid the critical decision thresholds. Consequently, there is ample
room to continuously deform these limits without altering the final logical
mapping---a topological flexibility that will be crucial for demonstrating
the structural stability of the set of solutions, as discussed later.

In Figs. \ref{graph1} and \ref{graph2}, the invariant zone $I_{0}$ is
represented by the sky-blue strip, while the invariant zone $I_{1}$ is
represented by the light-coral strip. The regions excluded by (\ref{outbas})
are visually represented by the light-grey strip surrounding the output
boundary (\ref{output}).

The four key logical regions $A_{ij}$, $i,j\in \{0,1\}$, representing the
exact discrete states of the embedded finite-state machine, are defined by
the Cartesian products of the memory basins and the readout domains in the
latent space as follows: 
\begin{equation}
A_{ij}\equiv F^{-1}(I_{i}\times J_{j})\cap \lbrack -1,1]^{d}\qquad \text{%
(Carry }i\text{, Output }j\text{),}  \label{invzones}
\end{equation}%
where $F:h\mapsto (Z,Y)$ is the linear mapping from the latent space to the
joint space formed by the scalar memory compression variable $Z$ and the
linear readout logit $Y$.

The pictures show that the entire sky-blue strip $I_{0}$ with tentative
carry 0 is mapped by $h_{t}(S_{t},h_{t-1})$ into the solid red, blue, and
green arcs for inputs $S_{t}=0$, $S_{t}=1$, and $S_{t}=2$, respectively. The
entire light-coral strip $I_{1}$ with tentative carry 1 is mapped into the
dashed red, blue, and green arcs for inputs $S_{t}=0$, $S_{t}=1$, and $%
S_{t}=2$, respectively.

In particular, all arcs are strictly contained within the interiors of the
strips $I_{0}$ and $I_{1}$. Since all points within the strips are mapped as
described, at the next time step the arcs themselves will also be mapped
accordingly, inheriting the rules from the strips they belong to. 

We now demonstrate zero-error infinite stability by verifying all possible
mappings one by one.

$i$) The solid red arc (Carry 0, Output 0) is located in the sky-blue zone $%
A_{00}$. Hence, it is mapped into itself for $S_{t}=0$, into the solid blue
arc (sky-blue zone $A_{01}$, Carry 0, Output 1) for $S_{t}=1$, and into the
solid green arc (light-coral zone $A_{10}$, Carry 1, Output 0) for $S_{t}=2$.

$ii$) The solid blue arc (Carry 0, Output 1) is located in the sky-blue zone 
$A_{01}$. Hence, it is mapped into the solid red arc (sky-blue zone $A_{00}$%
, Carry 0, Output 0) for $S_{t}=0$, into itself for $S_{t}=1$, and into the
solid green arc (light-coral zone $A_{10}$, Carry 1, Output 0) for $S_{t}=2$.

$iii$) The solid green arc (Carry 1, Output 0) lies in the light-coral zone $%
A_{10}$, so it is mapped into the dashed red arc (sky-blue zone $A_{01}$,
Carry 0, Output 1) for $S_{t}=0$. For $S_{t}=1$ it is mapped into the dashed
blue arc (light-coral zone $A_{10}$, Carry 1, Output 0), while it is mapped
into the dashed green arc (light-coral zone $A_{11}$, Carry 1, Output 1) for 
$S_{t}=2$.

This concludes the mapping of the solid arcs. Now we extend the analysis to
the dashed ones.

$iv$) The dashed red arc (Carry 0, Output 1) lies in the sky-blue zone $%
A_{01}$, thus it is mapped into the solid red arc (sky-blue zone $A_{00}$,
Carry 0, Output 0) for $S_{t}=0$, into the solid blue arc (sky-blue zone $%
A_{01}$, Carry 0, Output 1) for $S_{t}=1$, and into the solid green arc
(light-coral zone $A_{10}$, Carry 1, Output 0) for $S_{t}=2$.

$v$) The dashed blue arc (Carry 1, Output 0) lies in the light-coral zone $%
A_{10}$ and is thus mapped into the dashed red arc (sky-blue zone $A_{01}$,
Carry 0, Output 1) for $S_{t}=0$. For $S_{t}=1$ it is mapped into itself,
while for $S_{t}=2$ it is mapped into the dashed green arc (light-coral zone 
$A_{11}$, Carry 1, Output 1).

$vi$) The dashed green arc (Carry 1, Output 1) lies in the light-coral zone $%
A_{11}$, so it is mapped into the dashed red arc (sky-blue zone $A_{01}$,
Carry 0, Output 1) for $S_{t}=0$, into the dashed blue arc (light-coral zone 
$A_{10}$, Carry 1, Output 0) for $S_{t}=1$, and into itself for $S_{t}=2$.

All these mappings are in perfect agreement with logical binary addition (%
\ref{targetlogic})-(\ref{upcarry}). This proves zero-error infinite
stability.

Furthermore, as visually evident from the graphical analysis, the projected
arcs reside strictly within the interiors of their respective logical zones $%
A_{ij}$ (\ref{invzones}). This is crucial because it implies that the
specific infinitely stable system (\ref{system}) reproducing Peano
arithmetic with zero error is not unique. If we perturb the weight
parameters within a neighborhood of the values used above, the mathematical
proof remains valid, relying on the same sets $I_{i}$, $J_{j}$ and $A_{ij}$
in latent space. We give more details on this important property in the next
section.

Ultimately, the exact algebraic configuration (\ref{system}) is not an
isolated mathematical anomaly, but rather part of an open subset of exact
solutions within the parameter space. Because this subset is topologically
open, it inherently encompasses a dense set of configurations made of dyadic
rational numbers, reachable by the RNN.

These facts provide the fundamental reason why a neural network, trained
through optimized, iterative gradient descent and relying on truncated
floating-point arithmetic, can converge on a perfect solution. The
optimization algorithm is not forced to locate a dimensionless point in a
high-dimensional space; instead, it naturally descends into a broad,
error-tolerant topological basin that physiologically absorbs the
approximations of discrete computation.

\section{Analytical Proof of Zero-Error Topological Invariance in 
\texorpdfstring{$d=2$}{d=2}}

\label{sec:invariance}

At this point, it is straightforward to analytically prove that the learned
logic is stable with zero error within the continuous phase space, by
translating the graphical proof of the previous section into a detailed
analysis of the recurrent mapping $Z^{\prime }(S,Z)$ and the linear readout $%
Y(S,Z)$, which are 
\begin{equation}
Z^{\prime }(S,Z)=-\frac{2}{5}h_{1}(S,Z)-\frac{1}{4}h_{2}(S,Z),\qquad
Y(S,Z)=4h_{1}(S,Z)+h_{2}(S,Z)+1,  \label{eq:c_next}
\end{equation}%
where%
\begin{equation*}
h_{1}(S,Z)=\tanh \left( \frac{1}{4}S+Z-\frac{7}{10}\right) ,\qquad
h_{2}(S,Z)=\tanh \left( -\frac{9}{4}S-\frac{35}{4}Z+\frac{7}{2}\right) .
\end{equation*}

The transformation $(S,Z)\mapsto (h_{1}(S,Z),h_{2}(S,Z))$ with $S\in
\{0,1,2\}$ and $Z\in I_{0}\cup I_{1}$ gives the six arcs $a_{0S}$ and $%
a_{1S} $ shown in Fig. \ref{graph1}. We must prove that they are strictly
contained within the required logical regions $A_{ij}$, i.e.,%
\begin{eqnarray*}
&&\left[ 
\begin{tabular}{l}
$Z^{\prime }(0,I_{0})\Subset I_{0}$ \\ 
$Y(0,I_{0})\Subset J_{0}$%
\end{tabular}%
\right] ,\qquad \left[ 
\begin{tabular}{l}
$Z^{\prime }(1,I_{0})\Subset I_{0}$ \\ 
$Y(1,I_{0})\Subset J_{1}$%
\end{tabular}%
\right] ,\qquad \left[ 
\begin{tabular}{l}
$Z^{\prime }(2,I_{0})\Subset I_{1}$ \\ 
$Y(2,I_{0})\Subset J_{0}$%
\end{tabular}%
\right] , \\
&&\left[ 
\begin{tabular}{l}
$Z^{\prime }(0,I_{1})\Subset I_{0}$ \\ 
$Y(0,I_{1})\Subset J_{1}$%
\end{tabular}%
\right] ,\qquad \left[ 
\begin{tabular}{l}
$Z^{\prime }(1,I_{1})\Subset I_{1}$ \\ 
$Y(1,I_{1})\Subset J_{0}$%
\end{tabular}%
\right] ,\qquad \left[ 
\begin{tabular}{l}
$Z^{\prime }(2,I_{1})\Subset I_{1}$ \\ 
$Y(2,I_{1})\Subset J_{1}$%
\end{tabular}%
\right] ,
\end{eqnarray*}%
where the notation \textquotedblleft $A\Subset B$\textquotedblright\ means
that $A$ is contained within the interior of $B$, avoiding the boundaries.

Since formulas~(\ref{eq:c_next}) involve linear combinations of
monotonically increasing functions with opposing signs, local extrema may
exist within the intervals. Therefore, mapping only the interval boundaries
is insufficient to guarantee containment. By the Weierstrass Extreme Value
Theorem, the absolute bounds of the image of a closed interval under a
continuous function occur either at the boundaries or at the internal
critical points where the first derivative vanishes.

By calculating the extrema for $Z\in I_{0}$ and $Z\in I_{1}$, we obtain the
bounding boxes for all possible state transitions.

Transitions from $I_{0}$: for $S=0$, $Z^{\prime }\in \lbrack
-0.0115,0.0061]\Subset I_{0}$, with $Y\leq -0.3814<-0.3$; for $S=1$, $%
Z^{\prime }\in \lbrack -0.0487,-0.03799]\Subset I_{0}$, with $Y\geq
0.0687>0.03$; for $S=2$, $Z^{\prime }\in \lbrack 0.2150,0.2760]\Subset I_{1}$%
, with $Y\leq -0.4659<-0.3$.

Transitions from $I_{1}$: for $S=0$, $Z^{\prime }\in \lbrack
-0.0510,-0.0351]\Subset I_{0}$, with $Y\geq 0.1137>0.03$; for $S=1$, $%
Z^{\prime }\in \lbrack 0.2262,0.2766]\Subset I_{1}$, with $Y\leq
-0.4701<-0.3 $; for $S=2$, $Z^{\prime }\in \lbrack 0.2164,0.2443]\Subset
I_{1}$, with $Y\geq 0.0468>0.03$.

In all cases, both the mapped compression variable $Z^{\prime }$ and the
readout logit $Y$ are strictly confined within the interiors of their
respective target domains $I_{i}$ and $J_{j}$ with finite safety margins.

The dynamic state evolution under the continuous map $h=(h_{1},h_{2})$ in
the 2D latent phase space carries the four main logical regions $A_{ij}$, $%
i,j\in \{0,1\}$, into twelve image domains, which we denote as $H(S,A_{ij})$%
, $S\in \{0,1,2\}$. Because $A_{00}$ and $A_{01}$ are contained in the $%
I_{0} $ strip, while $A_{10}$ and $A_{11}$ are contained in the $I_{1}$
strip, the results just derived imply the $A_{ij}$ mappings 
\begin{equation}
H(S,A_{ij})\Subset A_{\lfloor (i+S)/2\rfloor ,\,(i+S)\,\mathrm{mod}2},
\label{containment}
\end{equation}%
or, more explicitly, 
\begin{eqnarray}
H(0,A_{00}) &\Subset &A_{00},\qquad H(1,A_{00})\Subset A_{01},\qquad
H(2,A_{00})\Subset A_{10},  \notag \\
H(0,A_{01}) &\Subset &A_{00},\qquad H(1,A_{01})\Subset A_{01},\qquad
H(2,A_{01})\Subset A_{10},  \notag \\
H(0,A_{10}) &\Subset &A_{01},\qquad H(1,A_{10})\Subset A_{10},\qquad
H(2,A_{10})\Subset A_{11},  \notag \\
H(0,A_{11}) &\Subset &A_{01},\qquad H(1,A_{11})\Subset A_{10},\qquad
H(2,A_{11})\Subset A_{11},  \label{12}
\end{eqnarray}%
in perfect agreement with Peano arithmetic. At the next time step, the
images are mapped according to the $A_{ij}$ region they inhabit, repeating
this cycle ad infinitum and guaranteeing absolute zero-error stability.

From a geometric perspective, in the specific rank-1 model (\ref{solution})
the scalar compression from 2D to 1D pairwise collapses the images $%
H(S,A_{ij})$ into subsets of the $6$ 1D arcs $a_{0S}$ and $a_{1S}$. Because $%
A_{00}$ and $A_{01}$ share the same memory basin $I_{0}$, while $A_{10}$ and 
$A_{11}$ share $I_{1}$, we have: 
\begin{equation}
H(S,A_{00})\subseteq a_{0S},\qquad H(S,A_{01})\subseteq a_{0S},\qquad
H(S,A_{10})\subseteq \text{ }a_{1S},\qquad H(S,A_{11})\subseteq \text{ }%
a_{1S}.  \label{scoll}
\end{equation}

In more general architectures, such as the non-compressed models discussed
in Appendix \ref{A}, the full rank of $W_{hh}$ prevents the scalar collapse (%
\ref{scoll}), leaving the twelve image regions $H(S,A_{ij})$ as distinct 2D
subdomains.

Specifically, when the network parameters $w=(W,b)$ from (\ref{solution})
are subjected to a small variation $\delta w$ while holding the composite
arrival zones $A_{ij}$ fixed in the latent space, the output decision
boundary $y_{t}=0$ possibly shifts and tilts, and if the perturbed recurrent
matrix $W_{hh}^{\prime }$ loses its rank-1 structure, the exact 2D $%
\rightarrow $ 1D scalar compression is broken, causing the 6 arcs $a_{0S}$
and $a_{1S}$ to inflate into 12 extended regions $H(S,A_{ij})$.

Nevertheless, because the compact logical zones $A_{ij}$ are separated by
finite gaps and the parameter choice in (\ref{solution}) guarantees that the
original arcs $a_{0S}$ and $a_{1S}$ lie strictly within the $A_{ij}$
interiors, for any sufficiently small perturbation $|\delta w|<\epsilon $
the displaced output boundary is kept safely away from the $A_{ij}$ zones
and the images $H(S,A_{ij})$ remain strictly confined within the required $%
A_{kl}$ interiors, ensuring zero-error infinite stability.

This formally proves that the strict interior topological containment (\ref%
{12}) continues to hold across an open set of exact solutions. This set
inherently encompasses a dense subset of dyadic rational weights, ensuring
that infinitely many machine-representable solutions exist, rather than a
singular, fine-tuned parameter configuration. For this reason, an optimized
RNN, trained through the repulsive dead-zone loss function, readily
converges to one of them without requiring infinite precision.

Furthermore, the existence of an open set of exact solutions in the
parameter space fundamentally bridges continuous dynamics and the physical
limits of finite-precision hardware. Digital computers evaluate the
transcendental hyperbolic tangent function via discrete dyadic rational
approximations, effectively substituting the ideal continuous manifold with
a highly granular multidimensional grid. However, because the compact
arrival zones $A_{ij}$ are separated by finite gaps and the strict interior
containment (\ref{containment}) holds across their twelve mappings, the
dynamical system inherently absorbs these microscopic floating-point
truncation errors, as long as the perturbation induced by the hardware's
numerical approximation remains sufficiently smaller than both the gaps
between the $A_{ij}$ regions and the internal containment margins separating
the images $H(S,A_{ij})$ from the $A_{ij}$ boundaries. This structural
robustness guarantees that the infinite stability of the embedded
finite-state machine holds true not only in ideal continuous mathematics,
but also in physical finite-precision implementations.

\section{Exact Solutions in \texorpdfstring{$d=3$}{d=3} Latent Space}

\label{sec:exact_solution_d3}

While the solutions from the two-dimensional latent space can be trivially
embedded into higher dimensions, the geometric freedom provided by
additional latent components allows for the direct algebraic construction of
exact solutions, significantly simplifying the formal proofs of zero-error
infinite stability and topological separation.

As always, given the binary input vector $x_{t}=(x_{1,t},x_{2,t})^{T}\in
\{0,1\}^{2}$ with total sum $S_{t}=x_{1,t}+x_{2,t}\in \{0,1,2\}$, the state
vector $h_{t}=(h_{1,t},h_{2,t},h_{3,t})^{T}\in \lbrack -1,1]^{3}$ evolves
according to (\ref{eq:general_hidden_update}) and the output (\ref{eq:output}%
) is determined by (\ref{eq:general_readout}).

\begin{figure}[tb]
\centering\includegraphics[width=16truecm]{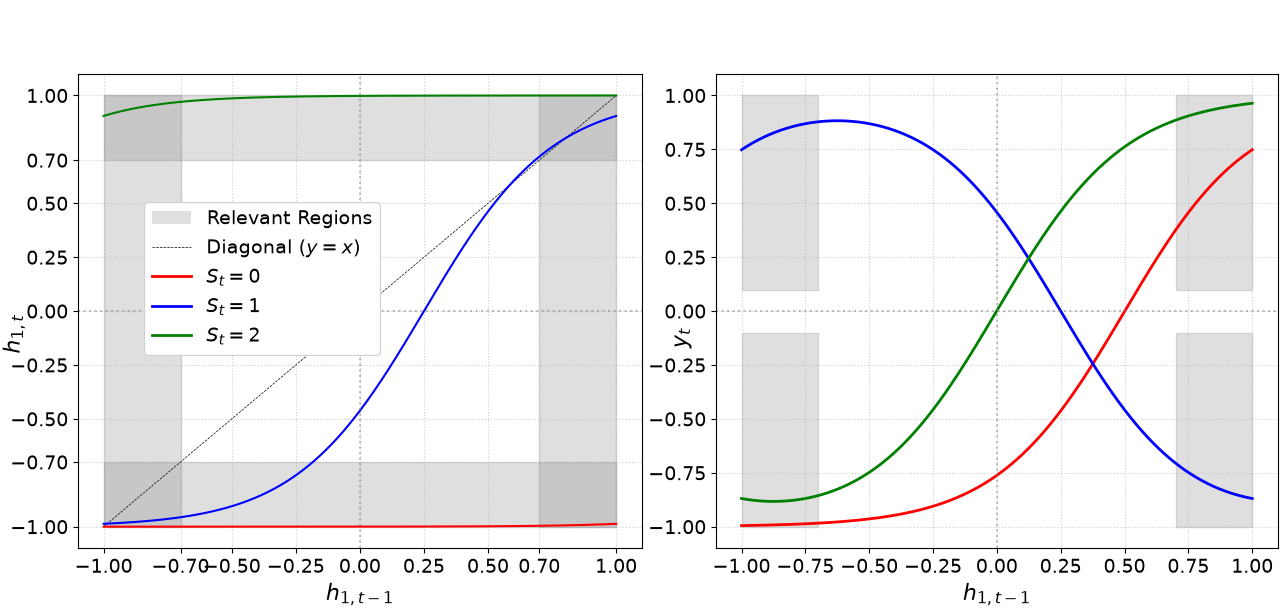}
\caption{Exact model\ (\protect\ref{sol3D}) in three latent space
dimensions. Carry (left picture) and Output (right picture)}
\label{grafico_1}
\end{figure}

The first system we consider is 
\begin{eqnarray}
h_{1,t} &=&\tanh (4S_{t}+2h_{1,t-1}-4.5),\qquad
y_{t}=-h_{1,t}+h_{2,t}+h_{3,t},  \notag \\
h_{2,t} &=&\tanh (4S_{t}+2h_{1,t-1}-1),\qquad \quad \hat{y}_{t}=\frac{1}{2}%
\left( \text{sign}(y_{t})+1\right) ,  \notag \\
h_{3,t} &=&\tanh (4S_{t}+2h_{1,t-1}-8).\qquad  \label{sol3D}
\end{eqnarray}

To prove that the carry and output evolve as required, we refer to Fig. \ref%
{grafico_1}. Taking $Z=h_{1}$ and $Y=-h_{1}+h_{2}+h_{3}$, we introduce the
candidate memory basins and readout ranges%
\begin{equation}
Z\in I_{0}=[-1,-0.7],\qquad Z\in I_{1}=[0.7,1],\qquad Y\in
J_{0}=[-3,-0.1],\qquad Y\in J_{1}=[0.1,3],  \label{basi}
\end{equation}%
which define the logical regions $A_{ij}$ as specified by Eq. (\ref{invzones}%
). As before, the intervals $I_{0}$ and $I_{1}$ correspond to Carry 0 and
Carry 1, respectively, while the intervals $J_{0}$ and $J_{1}$ correspond to
Binary Output 0 and Binary Output 1. The extremum $0.7$ of $I_{1}$ is to
some extent free, provided it remains larger than the solution $b^{\ast
}\approx 0.5851$ of the equation $b^{\ast }=\tanh (2b^{\ast }-0.5)$ (which
determines the point where $h_{1,t}$ for $S_{t}=1$ meets the diagonal), and
sufficiently large to yield the right readout logit, upon adjusting the
other entries compatibly.

It is evident from Fig. \ref{grafico_1} that if $h_{1,t-1}\in I_{0}$ (Carry
0), $S_{t}=0$ gives $h_{1,t}\in I_{0}$ and $y_{t}\in J_{0}$ (logical zone $%
A_{00}$, Carry 0, Output 0), $S_{t}=1$ gives $h_{1,t}\in I_{0}$ and $%
y_{t}\in J_{1}$ (logical zone $A_{01}$, Carry 0, Output 1), while $S_{t}=2$
gives $h_{1,t}\in I_{1}$ and $y_{t}\in J_{0}$ (logical zone $A_{10}$, Carry
1, Output 0).

On the other hand, if $h_{1,t-1}\in I_{1}$ (Carry 1), $S_{t}=0$ gives $%
h_{1,t}\in I_{0}$ and $y_{t}\in J_{1}$ (logical zone $A_{01}$, Carry 0,
Output 1), $S_{t}=1$ gives $h_{1,t}\in I_{1}$ and $y_{t}\in J_{0}$ (logical
zone $A_{10}$, Carry 1, Output 0), while $S_{t}=2$ gives $h_{1,t}\in I_{1}$
and $y_{t}\in J_{1}$ (logical zone $A_{11}$, Carry 1, Output 1).

The mapped state transitions and readout signs strictly embody the inductive
axioms of Peano arithmetic for binary addition. As in two latent-space
dimensions, the arrangement (\ref{sol3D}) guarantees zero-error infinite
stability.

Moreover, Fig. \ref{grafico_1} shows that the logical regions $A_{ij}$,
which are compact and separated by finite gaps, are strictly mapped by the
dynamics into their interiors, according to the containment formula (\ref%
{containment}), maintaining safety margins that enable robust topological
invariance. This ensures that the proof of infinite stability continues to
hold if we perturb the weights and biases within neighborhoods of the values
appearing in (\ref{sol3D}), confirming that (\ref{sol3D}) belongs to an open
set of zero-error exact solutions in parameter space. This makes it easily
reachable by an optimized RNN training.

\begin{figure}[tb]
\centering\includegraphics[width=16truecm]{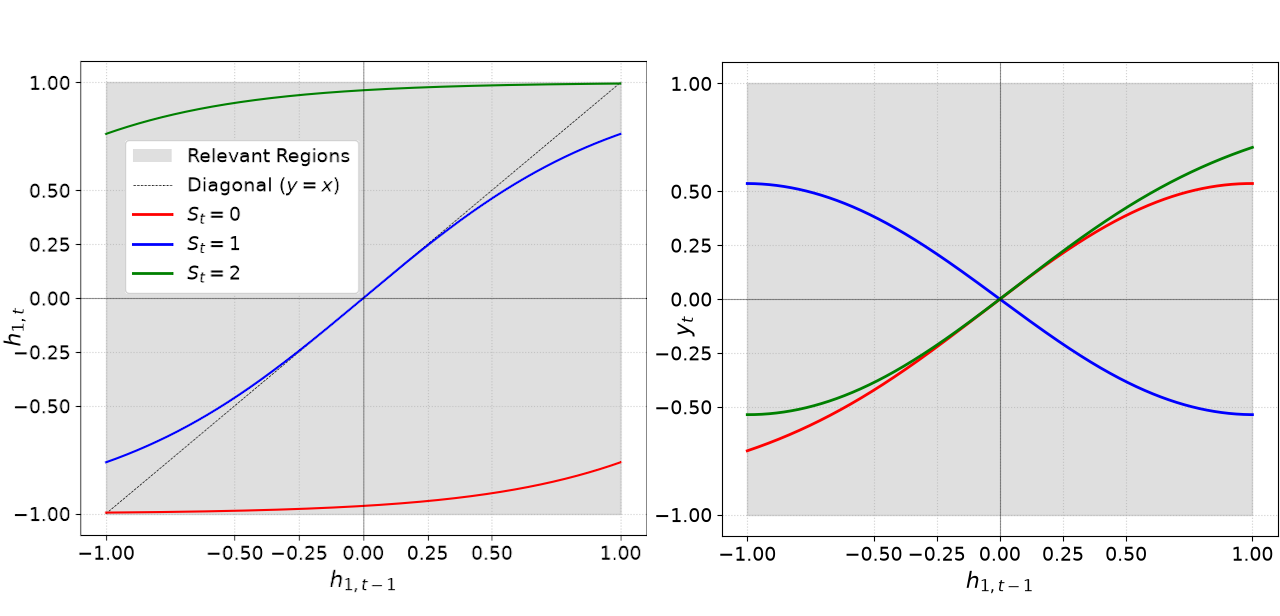}
\caption{Carry and output of the system (\protect\ref{scal})}
\label{grafico_2}
\end{figure}

At the same time, exact solutions that are not reachable on finite-precision
hardware (because, for example, they involve transcendental weights) are
straightforward to construct. One such configuration is%
\begin{eqnarray}
h_{1,t} &=&\tanh (2S_{t}+h_{1,t-1}-2),\qquad h_{2,t}=\tanh
(2S_{t}+h_{1,t-1}),  \notag \\
h_{3,t} &=&\tanh (2S_{t}+h_{1,t-1}-4),\qquad y_{t}=-\tanh (4)h_{1,t}+\tanh
(2)(h_{2,t}+h_{3,t}).  \label{scal}
\end{eqnarray}%
Referring to Fig. \ref{grafico_2}, the analysis proceeds as before, but now
the memory basins and readout ranges are given by $I_{0}=[-1,0)$, $%
I_{1}=(0,1]$, $J_{0}=[-\tanh (4)-2\tanh (2),0)$, $J_{1}=(0,\tanh (4)+2\tanh
(2)]$.

As established in Section \ref{sec:empirical_verification}, setting the
input to $S_{t}=1$ serves as a tentative approach for uncovering the
internal carry memory. By substituting $S_{t}=1$ into the evolution equation
(\ref{scal}) for $h_{1,t}$, the dynamics strictly reduces to $h_{1,t}=\tanh
(h_{1,t-1})$. Because the hyperbolic tangent preserves the sign of its
argument indefinitely, the state trajectory can never cross the origin.
Therefore, by using the sign of $h_{1,t-1}$ as a proxy for the logical
carry, the memory state remains on the same side of the carry boundary until
an input $S_{t}\in \{0,2\}$ drives a transition across, landing on the
correct side in all cases.

Building upon this shielded memory, the readout logit $y_{t}$ must
dynamically decode the binary addition. The transcendental coefficients of $%
y_{t}$ have been adjusted to guarantee that $y_{t}$ and $h_{1,t-1}$ share
the same sign when $S_{t}\in \{0,2\}$, and have strictly opposite signs when 
$S_{t}=1$. This precise geometric alignment ensures that the continuous
readout correctly implements the discrete Boolean logic of binary addition,
granting the continuous $d=3$ model zero-error stability at infinity.

The drawback of this model is that the gaps partitioning the invariant
memory basins and the readout regions are equal to zero. Under continuous
inputs of $S_{t}=1$, for example, the carry proxy decays according to $%
h_{1,t}=\tanh (h_{1,t-1})$, and its absolute value $|h_{1,t}|$ tends to 0.

On the other hand, in unscaled configurations, such as%
\begin{equation}
y_{t}=-h_{1,t}+h_{2,t}+h_{3,t},  \label{out2}
\end{equation}%
the output decision boundary $y_{t}=0$ exposes a structural misalignment.
Substituting the state equations into (\ref{out2}) yields the root $%
2S_{t}+h_{1,t-1}\approx 0.0381$, with $y_{t}$ being positive above this
value and negative below it. Under continuous neutral inputs $S_{t}=1$, the
carry proxy strictly decays towards zero, as noted above. While the memory
component $h_{1,t}$ preserves its original sign and never crosses the
origin, its strictly diminishing absolute value eventually causes the
combined argument $2S_{t}+h_{1,t-1}$ to fall below the misaligned threshold
of $0.0381$ at $S_{t}=0$. So, if a long list of inputs $S_{t}=1$ is followed
by input $S_{t}=0$, this threshold is crossed, and the output logit $y_{t}$
prematurely inverts its sign, triggering a topological breakdown where the
structurally preserved internal memory is incorrectly decoded by the readout
layer.

It is possible to establish a finite gap between the readout regions $A_{ij}$%
, ensuring that their images lie strictly inside their own interiors, by
twisting the weights as follows:%
\begin{eqnarray}
h_{1,t} &=&\tanh (2S_{t}+1.3h_{1,t-1}-2.03),\qquad h_{2,t}=\tanh
(2S_{t}+1.3h_{1,t-1}-0.03),\quad  \notag \\
h_{3,t} &=&\tanh (2S_{t}+1.3h_{1,t-1}-4),\qquad
y_{t}=-h_{1,t}+h_{2,t}+h_{3,t}.  \label{3Dtwist}
\end{eqnarray}%
As before, we take $Z=h_{1}$ and $Y=-h_{1}+h_{2}+h_{3}$. We define the
invariant memory sets and readout intervals as%
\begin{equation}
Z\in I_{0}=[-1,-0.25],\quad Z\in I_{1}=[0.25,1],\quad Y\in
J_{0}=[-3,-0.1],\quad Y\in J_{1}=[0.1,3],
\end{equation}%
where the entry $0.25$ of $I_{1}$ is a number $b>b^{\ast }\approx 0.1012$
(solution of the equation $b^{\ast }=\tanh (1.3b^{\ast }-0.03)$). This is to
have the right carry propagation and readout logit, upon adjusting the other
entries accordingly.

\begin{figure}[tb]
\centering\includegraphics[width=16truecm]{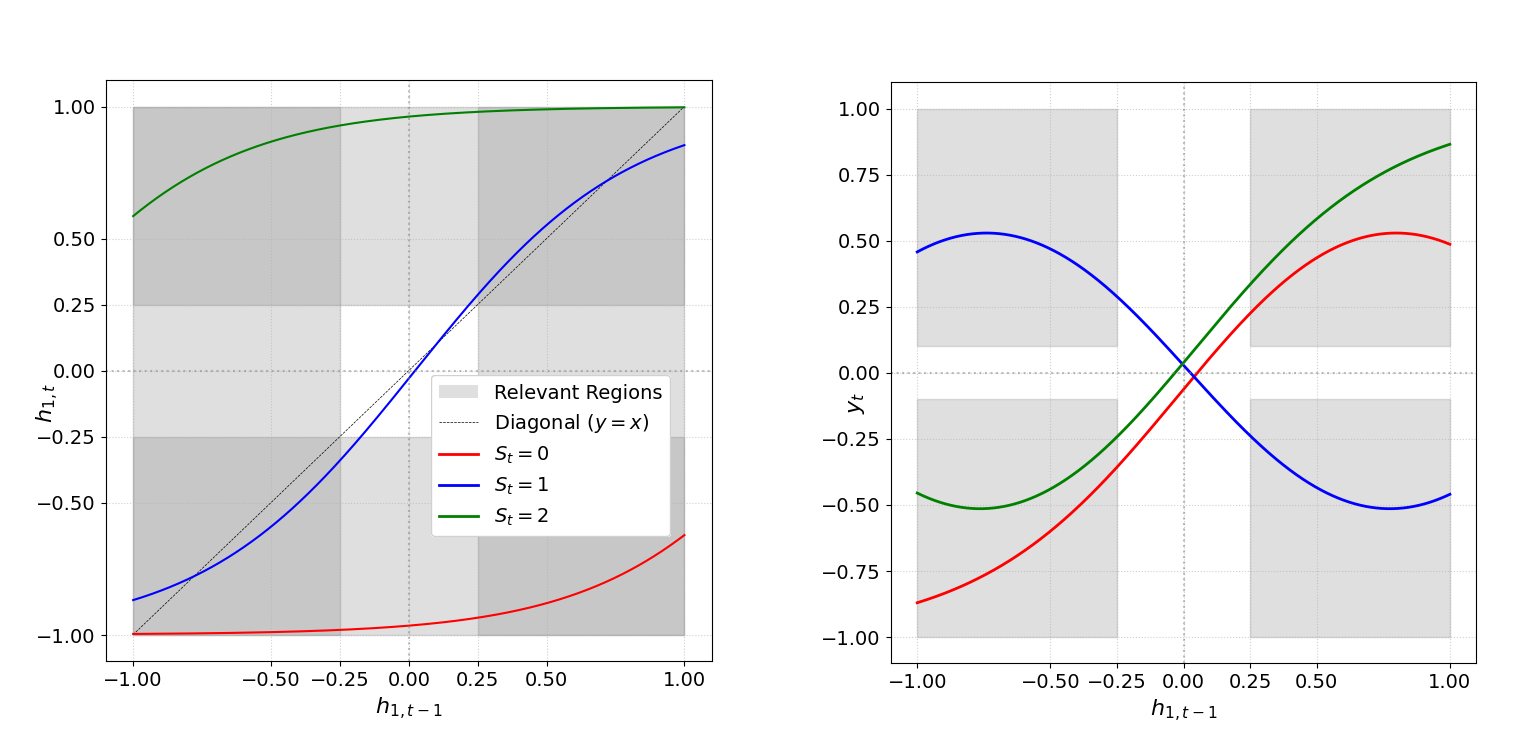}
\caption{Carry and output of the system (\protect\ref{3Dtwist})}
\label{grafico_3}
\end{figure}

The proof of zero-error infinite stability within the logical zones $A_{ij}$%
, defined in Eq. (\ref{invzones}), proceeds as before, and can be checked
through Fig. \ref{grafico_3}. The analysis also confirms that the
configuration (\ref{3Dtwist}) belongs to an open set of zero-error exact
solutions in parameter space.

An interesting theoretical possibility is to replace the hyperbolic tangent
with the sign function. While this substitution collapses the continuous
phase space onto exact, immobile fixed points located at the vertices $%
h_{t}\in \{-1,+1\}^{3}$ of a 3D hypercube, it is fundamentally unsuitable
for standard RNN training via gradient descent, because the sign function
possesses a derivative of zero almost everywhere and an undefined gradient
at the point of discontinuity. The backpropagation algorithm is stripped of
the continuous scalar field required to guide weight updates, causing the
optimization process to immediately stagnate.

We retain the weight matrices of (\ref{scal}), apart from the output, which
no longer needs to be the one of (\ref{scal}) for infinite stability, and
can be replaced by (\ref{out2}). Defining the unified activation argument $%
A_{t}=2S_{t}+h_{1,t-1}$, the state equations and the readout logit decouple
into the following discrete operators: 
\begin{equation}
h_{1,t}=\text{sign}(A_{t}-2),\qquad h_{2,t}=\text{sign}(A_{t}),\qquad
h_{3,t}=\text{sign}(A_{t}-4),\qquad y_{t}=-h_{1,t}+h_{2,t}+h_{3,t}.
\label{eq:sign_h1}
\end{equation}

The carry $c_{t-1}$ equals 0 for $h_{1,t-1}=-1$ and 1 for $h_{1,t-1}=1$.
Since $S_{t}\in \{0,1,2\}$, $A_{t}$ is constrained to the set of odd
integers $\{-1,1,3,5\}$. The argument of the sign function is never zero,
eliminating any mathematical indeterminacy at the decision boundaries.

It is easy to verify that the system correctly executes all six logical
transitions of the binary adder, by exclusively visiting four distinct fixed
points in the three-dimensional latent space: $(-1,-1,-1)$, $(-1,1,-1)$, $%
(1,1,-1)$, and $(1,1,1)$.

\section{Exact Solutions in \texorpdfstring{$d=4$}{d=4} Latent Space}

\label{sec:exact_solution_d4}

In a four-dimensional phase space $%
h_{t}=(h_{1,t},h_{2,t},h_{3,t},h_{4,t})^{T}\in \lbrack -1,1]^{4}$, we first
consider the model whose four-component recurrent state evolution, as given
by Eq. (\ref{eq:general_hidden_update}), reads 
\begin{eqnarray}
h_{1,t} &=&\tanh (2S_{t}+h_{1,t-1}-2),\qquad h_{2,t}=\tanh
(2S_{t}-h_{1,t-1}-2),  \notag \\
h_{3,t} &=&\tanh (8S_{t}+h_{1,t-1}-8),\qquad h_{4,t}=\tanh
(8S_{t}-h_{1,t-1}-8),  \label{h4}
\end{eqnarray}%
and the linear readout layer of Eq. (\ref{eq:general_readout}) takes the form%
\begin{equation}
y_{1,t}=0,\qquad y_{2,t}=h_{1,t}-h_{2,t}-2h_{3,t}+2h_{4,t},  \label{y4}
\end{equation}%
yielding the predicted output bit $\hat{y}_{t}=[$sign$(y_{2,t})+1]/2$, by
formula (\ref{eq:output}). As usual, $x_{t}=(x_{1,t},x_{2,t})^{T}\in
\{0,1\}^{2}$ denotes the binary input vector at clock cycle $t$, and $%
S_{t}=x_{1,t}+x_{2,t}\in \{0,1,2\}$ is the total input sum.

As in the $d=3$ examples of the previous section, the first component $%
h_{1,t}$ serves as an exact continuous proxy for the logical carry bit: $%
c_{t}=0$ if $h_{1,t}<0$ and $c_{t}=1$ if $h_{1,t}>0$. We already know that
the carry propagates correctly, because $h_{1,t}(S_{t},h_{1,t-1})$ coincides
with the one of (\ref{scal}). It is immediate to show that the output $\hat{y%
}_{t}$ agrees with the target operation (\ref{targetlogic}), by evaluating
the readout logit as a function of the prior carry $c_{t-1}$. This
establishes the exact symbolic equivalence between continuous $\tanh $
updates and Peano arithmetic.

Nevertheless, the invariant basins lack a finite topological gap, and the
readout regions similarly afford zero safety margin. Consequently, the
configuration defined by (\ref{h4}) and (\ref{y4}) is merely marginally
stable and constitutes a fragile, fine-tuned state effectively unreachable
by gradient descent. This case is similar to the one of model (\ref{scal})
in $d=3$, with the difference that it does not involve transcendental
weights. 
\begin{figure}[tb]
\centering\includegraphics[width=16truecm]{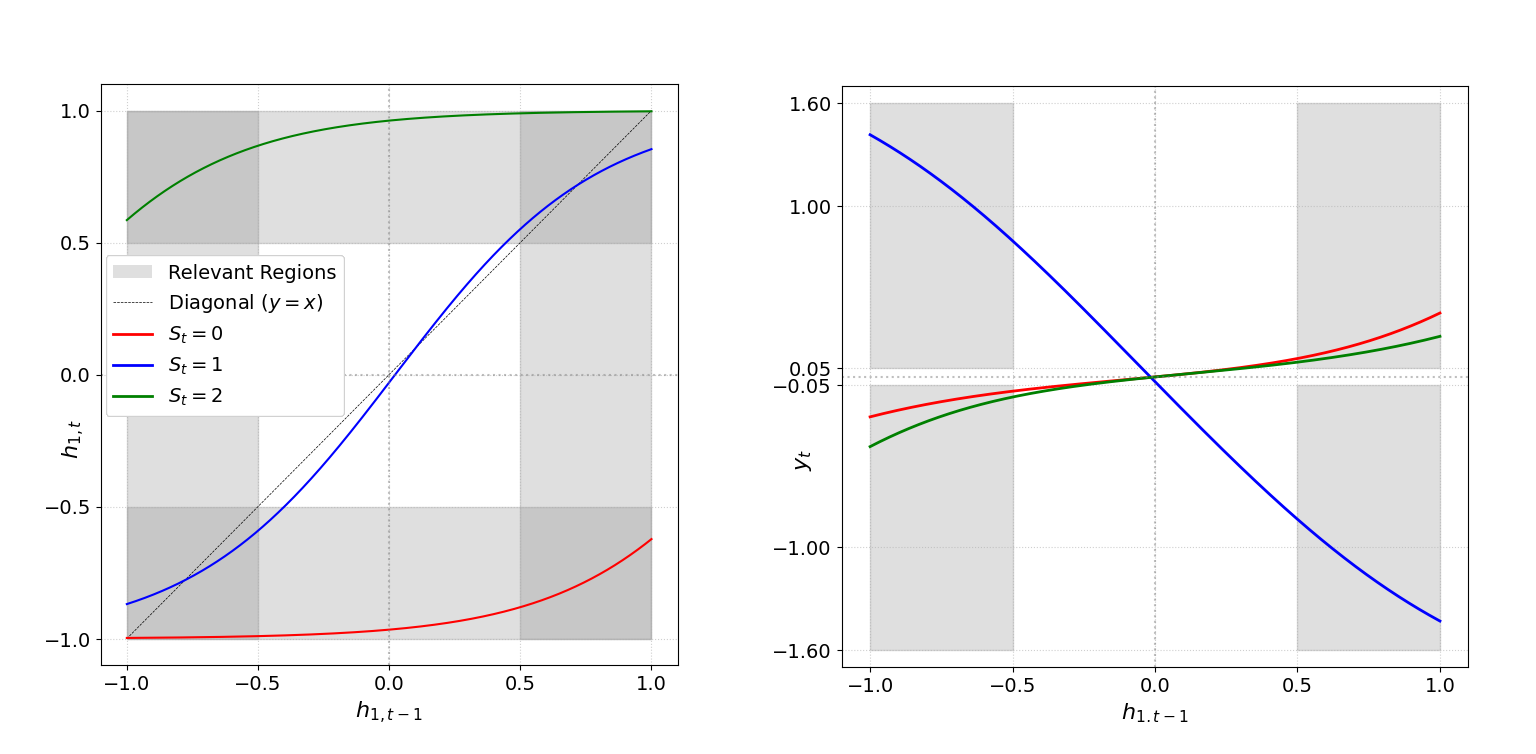}
\caption{Carry and output of the system (\protect\ref{h4})-(\protect\ref{y4}%
) with correction (\protect\ref{corr4})}
\label{grafico_4}
\end{figure}

We can achieve topological separation by applying a small structural
perturbation to the parameters. For example, it is sufficient to replace the
update rule for $h_{1,t}$ with the one of (\ref{3Dtwist}),%
\begin{equation}
h_{1,t}=\tanh (2S_{t}+1.3h_{1,t-1}-2.03),  \label{corr4}
\end{equation}%
and leave all other weights and biases unmodified. We then define the
invariant memory basins and readout regions as%
\begin{equation*}
Z\in I_{0}=[-1,-0.5],\quad Z\in I_{1}=[0.5,1],\quad Y\in
J_{0}=[-6,-0.05],\quad Y\in J_{1}=[0.05,6],
\end{equation*}%
for $Z=h_{1}$ and $Y=h_{1}-h_{2}-2h_{3}+2h_{4}$, where the lower bound $0.5$
of $I_{1}$ is adjusted to guarantee the correct carry propagation and output
logit. This value must be sufficiently larger than the root $a^{\ast
}\approx 0.1012$ of the equation $a^{\ast }=\tanh (1.3a^{\ast }-0.03)$, yet
appropriately bounded to maintain compatibility with the remaining
parameters.

The proof of zero-error infinite stability mirrors the previous analysis and
is illustrated by the phase space mapping visible in Fig. \ref{grafico_4}.

Besides achieving finite topological separation for both the invariant
basins and the compact logical zones $A_{ij}$, defined in Eq. (\ref{invzones}%
), Fig. \ref{grafico_4} shows that the dynamics in latent space maps each $%
A_{ij}$ into the interior of the required $A_{kl}$, according to the
containment formula (\ref{containment}). This proves that the exact solution
just outlined belongs to an open set of exact solutions, making it easily
reachable by optimized RNN training.

\section{Extension to Binary Multiplication: Modular Neuro-Symbolic
Composability}

\label{sec:binary_multiplication_extension}

While continuous binary addition requires a constant memory footprint to
store the single-bit carry state $c_{t}\in \{0,1\}$, expanding algorithmic
execution to binary multiplication requires accumulating partial products
across $2N$ clock cycles for $N$-bit operands. To prevent the network from
overwriting what it has already perfectly learned (a phenomenon known as 
\textit{catastrophic forgetting}) and to avoid a geometrically unmanageable
explosion of states (the curse of dimensionality) that would force it to
re-learn fundamental arithmetic from scratch, we decouple the computational
tasks by introducing a hybrid neuro-symbolic architecture. Specifically, we
separate the system into a frozen exact adder (the Arithmetic Logic Unit)
that performs the sums, and a lightweight neural controller trained
exclusively to manage temporal control flow, conditional gating, and spatial
shifting. This compositional approach mathematically mirrors human cognitive
development, where complex operations like multiplication are built upon the
previously mastered foundation of addition. By composing independently
stable tasks in sequence---rather than forcing a single network to
approximate the entire algorithmic mapping end-to-end---this framework
introduces a crucial vertical structure into artificial intelligence.
Consequently, it transforms a flat, opaque black box into a transparent
hierarchical system, maximizing both computational efficiency and structural
interpretability.

\subsubsection*{Hybrid Architecture and Operational Mechanics}

Let $A=(a_{1},a_{2},\dots ,a_{N})$ be the multiplicand vector and $%
B=(b_{1},b_{2},\dots ,b_{N})$ be the multiplier sequence presented
bit-serially from the least significant bit (the rightmost one,
corresponding to the positional weight $2^{0}$) to the most significant bit
(the leftmost one, associated with the maximum positional weight). The
multiplication task decomposes into the exact recurrence: 
\begin{equation}
P_{t}=P_{t-1}+g_{t}\cdot (A\ll s_{t}).  \label{recur}
\end{equation}%
This equation formalizes the discrete hardware logic of the shift-and-add
algorithm. Its components are defined as follows:

\begin{enumerate}
\item $P_{t}$ is the partial product accumulator. It maintains the
continuously updated running sum of the multiplication process, dynamically
aggregating the newly generated terms at each clock cycle, initialized at
the null state $P_{0}=0$.

\item $(A\ll s_{t})$ is the positionally shifted multiplicand, where the $%
\ll $ symbol (Bitwise Left Shift) represents the logical left shift
operator, which applies a binary left shift of $s_{t}\in \{0,1\}$ positions
to the input vector $A$, ensuring that the term is accurately aligned with
the corresponding positional power of 2 before being integrated into the
accumulator.

\item $g_{t}\in \{0,1\}$ is the logical gate decision. It acts as a
conditional switch corresponding to the current bit of the multiplier
sequence $B$. If $g_{t}=1$, the shifted multiplicand is added to the
accumulator; if $g_{t}=0$, the accumulator state remains invariant.
\end{enumerate}

Because the network must linearly decode two distinct operational logits,
the minimal architectural latent space requires two dimensions, $h_{t}\in
\lbrack -1,1]^{2}$, even if the optimized dynamics ultimately prune excess
degrees of freedom.

The hybrid neuro-symbolic system decouples execution into two specialized
functional units:

\begin{description}
\item[a)] Frozen Neural ALU (Exact Adder): The pre-trained RNN adder
preserves its weight matrices ($W_{ih},W_{hh},W_{fc}$) and bias vectors ($%
b_{h},b_{fc}$), executing exact binary additions with carry propagation
without parameter updates.

\item[b)] Trainable Neural Controller: A minimal recurrent controller
network receives as input the bit $x_{t}\in \{0,1\}$ belonging to the
multiplier $B$ at step $t$ and updates its internal latent state: 
\begin{equation*}
h_{t}=\tanh (W_{ih}x_{t}+W_{hh}h_{t-1}+b_{h}).
\end{equation*}
This state is then mapped to operational outputs via a linear readout layer,
emitting two logits whose zero-level sets establish strict secant lines
partitioning the 2D latent phase space: 
\begin{equation}
y_{t}=W_{fc}h_{t}+b_{fc}=\left( 
\begin{tabular}{l}
$y_{\text{gate},t}$ \\ 
$y_{\text{shift},t}$%
\end{tabular}%
\right) .  \label{sysmul}
\end{equation}
\end{description}

The network is trained on the lowest-valued arithmetic
operations---exhaustively covering the $16\times 16=256$ combinatorial
instances of numbers up to 4 bits---and subsequently subjected to stress
tests on ultra-long bit sequences (see below).

During the training phase, the components of the raw two-dimensional logit
vector $y_{t}$ are mapped to independent probability distributions. Using
sigmoid functions, we define%
\begin{equation}
p_{\text{gate},t}=\frac{1}{1+e^{-y_{\text{gate},t}}},\qquad p_{\text{shift}%
,t}=\frac{1}{1+e^{-y_{\text{shift},t}}},  \label{sigp}
\end{equation}%
as the network's estimated probabilities for executing the gating ($g_{t}=1$%
) and shifting ($s_{t}=1$) operations, respectively. These probabilities
yield values strictly greater than 1/2 for positive logits ($y_{\text{gate}%
,t}$, $y_{\text{shift},t}>0$) and strictly less than 1/2 for negative logits
($y_{\text{gate},t}$, $y_{\text{shift},t}<0$).

The estimated probabilities (\ref{sigp}) are evaluated against the exact
binary targets read from the dataset\ in order to extract $p_{t}$, the
probability assigned to the correct logical class. For the conditional gate, 
$p_{t}$ dynamically evaluates to $p_{\text{gate},t}$ when the dataset target
is 1, and to its complement $1-p_{\text{gate},t}$ when the target is 0. For
the unconditional shift, the target is always 1, so $p_{t}$ strictly
evaluates to $p_{\text{shift},t}$.

These $p_{t}$ variables are then fed independently into the repulsive
dead-zone loss function $\mathcal{L}_{\text{RDZ}}(p_{t})$ defined in Eq. (%
\ref{eq:dead_zone_loss}). If the network's prediction lacks sufficient
confidence, the logarithmic penalty generates the non-zero gradients
required to update the weights.

Once the network has finished learning and is deployed for operational
inference, it makes its logical decisions based solely on whether the raw
linear value (the logit) is above or below zero, according to the formulas 
\begin{equation*}
g_{t}=\frac{1}{2}[\text{sign}(y_{\text{gate},t})+1]\quad \text{and}\quad
s_{t}=\frac{1}{2}[\text{sign}(y_{\text{shift},t})+1].
\end{equation*}%
This hard-thresholding mechanism is perfectly consistent with the
probabilistic interpretation introduced in Eq. (\ref{sigp}). For the
conditional gate, a positive logit $y_{\text{gate},t}>0$ ($g_{t}=1$) implies 
$p_{\text{gate},t}>1-p_{\text{gate},t}$, thereby selecting a target of $1$.
Conversely, a negative logit $y_{\text{gate},t}<0$ ($g_{t}=0$) corresponds
to $p_{\text{gate},t}<1-p_{\text{gate},t}$, thereby selecting a target of $0$%
. For the unconditional shift operation, the network's optimization must
guarantee $y_{\text{shift},t}>0$ across all valid trajectories; this ensures 
$p_{\text{shift},t}>1-p_{\text{shift},t}$ at all times, thereby consistently
locking the decision to the required target $s_{t}=1$ at every clock cycle.

We take a moment to explain the fundamental logical difference between the
tasks of arithmetic addition and multiplication. While both modules share a
linear readout layer to generate the output logits, their training
objectives require different activation functions to map the raw logits $%
y_{t}$ into valid probabilities for the repulsive dead-zone loss.

\begin{itemize}
\item \textbf{Addition (Mutually Exclusive Logic):} The binary adder must
predict a single output bit (either 0 or 1). Because these states are
mutually exclusive for a single variable, the network outputs a
two-dimensional logit vector that is normalized via a softmax function (\ref%
{prediction}). This couples the outputs, ensuring that their probabilities
sum to one.

\item \textbf{Multiplication (Independent Logic):} The multiplication
controller must make two functionally independent binary decisions
simultaneously: whether to Gate (accumulate) and whether to Shift (translate
spatial coordinates). Because these actions do not exclude each other (the
network must frequently shift without gating), their respective logits
cannot be coupled into a single probability distribution. Instead, they must
be evaluated using the independent sigmoid functions (\ref{sigp}).
\end{itemize}

\subsubsection*{RNN\ Solution and Stress Test}

One of the many systems (\ref{sysmul}) generated by the RNN training (in
less than 60 epochs)\ is 
\begin{align}
h_{1,t}& =\tanh (0.8278\cdot x_{t}+1.1571\cdot h_{1,t-1}-0.5827\cdot
h_{2,t-1}+1.2371),  \notag \\
h_{2,t}& =\tanh (-2.2772\cdot x_{t}+0.1851\cdot h_{1,t-1}-0.0047\cdot
h_{2,t-1}+0.7356),  \notag \\
y_{\text{gate},t}& =0.5266\cdot h_{1,t}-2.8921\cdot h_{2,t}-0.8551,  \notag
\\
y_{\text{shift},t}& =1.9006\cdot h_{1,t}-0.7983\cdot h_{2,t}+1.0950.
\label{solmult}
\end{align}%
For simplicity, we truncated the numbers to four decimal figures, without
affecting the properties stated below.

This solution was probed by means of stress tests that, rather than
performing standard finite-length arithmetic operations, isolated the
dynamic component and constrained it to navigate its continuous phase space
over an unbroken sequence of clock cycles. For each temporal length $L\in
\{2^{6},2^{10},2^{12},2^{16},2^{18},2^{20}\}$, the system was evaluated
without resetting the latent state $h_{t}$, enforcing the continuous
accumulation of float32 truncation noise. At each step, the outputs were
verified against the strict shift-and-add Boolean invariants. The gate
decision perfectly matched the input bit ($g_{t}=x_{t}$), and the shift
command remained unconditionally active ($s_{t}=1$).

\subsubsection*{Analytical Proof of Controller Dynamics}

Now we provide a mathematical proof that the recurrent neural controller (%
\ref{solmult}) perpetually executes the correct shift-and-add logic without
failure. First, we show that the rectangle 
\begin{equation}
(h_{1,t-1},h_{2,t-1})\in \lbrack -1,1]\times \lbrack -1,0.73].
\label{invbas}
\end{equation}%
maps into itself, i.e., it is an invariant basin of the dynamical system.

Due to the monotonicity of the hyperbolic tangent, it is sufficient to check
the maximum and the minimum of $h_{2,t}$, which are evaluated at the
geometric extremes $h_{1,t-1}=1$, $h_{2,t-1}=-1$ and $h_{1,t-1}=-1$, $%
h_{2,t-1}=0.73$, respectively. For $x_{t}=0$ the updated state satisfies $%
h_{2,t}\in \lbrack 0.498,0.728]$, while for $x_{t}=1$ it satisfies $%
h_{2,t}\in \lbrack -0.939,-0.874]$. Both lie within (\ref{invbas}), so the
restricted basin is an invariant set, as claimed.

The $y_{\text{gate},t}$ readout is a linear combination of the latent
variables $h_{1,t}$ and $h_{2,t}$. Combining the unconditional domain $%
[-1,1] $ of $h_{1,t}$ with the conditional intervals of $h_{2,t}$
established above, we obtain the conservative bounds $y_{\text{gate},t}\leq
0.5266(1)-2.8921(0.498)-0.8551=-1.769$ for $x_{t}=0$ and $y_{\text{gate}%
,t}\geq 0.5266(-1)-2.8921(-0.874)-0.8551=1.146$ for $x_{t}=1$, perfectly
matching the input bit logic with a structural safety margin.

In the case of $y_{\text{shift},t}$, an analysis of this type is not
resolutive. For $x_{t}=1$, we refine it by evaluating the strict minimum of $%
h_{1,t}$ at the worst-case vertex $(-1,0.73)$ of (\ref{invbas}), which
yields $h_{1,t}\geq 0.448$. Substituting this gives $y_{\text{shift},t}\geq
1.9006(0.448)-0.7983(-0.874)+1.0950=2.644>0$.

For $x_{t}=0$, even this refinement is not resolutive. The inequality $%
\partial y_{\text{shift},t}/\partial h_{1,t-1}+2\partial y_{\text{shift}%
,t}/\partial h_{2,t-1}<0$, which is straightforward to prove, excludes local
maxima and minima in the interior of (\ref{invbas}). Studying the partial
derivatives on the edges of the rectangle, it is easy to prove that the
minimum of $y_{\text{shift},t}$ in (\ref{invbas}) occurs at the vertex $%
(-1,0.73)$ and evaluates to $0.0657>0$. The shift command is therefore
unconditionally active, completing the analytical proof of the algorithmic
routing.

This analysis formally proves that within the invariant set (\ref{invbas}),
the recurrent controller maintains strict, perpetual adherence to the target
algorithmic instructions, permanently shielded from numerical truncation
noise across an infinite horizon. Moreover, the safety margins identified
above ensure that the zero-error infinitely stable solutions include an open
set in parameter space, thereby allowing a suitably trained RNN\ to reach
one of them easily.

Ultimately, we have demonstrated that properly trained neural modules
through the repulsive dead-zone loss can be hierarchically composed to
construct arbitrarily complex, noise-immune algorithmic processors. It is
worth emphasizing that the network was never explicitly instructed to
execute the classical shift-and-add algorithm; rather, it was trained solely
on input-output pairings using unconstrained operational triggers for the
gating and shifting actions. Autonomously, it suppressed its temporal
memory, transforming itself from a memory-based dynamical system into a
pure, instantaneous logical switch.

By design, a Recurrent Neural Network accumulates memory over time,
generating complex trajectories within its latent space. Given its
continuous multidimensional capacity, the network could easily have
entangled its operations in spurious past states and superfluous hysteresis.
Instead, the geometric pressure exerted by the repulsive dead-zone loss
acted as an algorithmic Occam's razor, forcing the optimizer to eliminate
redundant degrees of freedom. The gradient descent process autonomously
pruned all unnecessary memory accumulation, strictly preserving only the
discrete logic required for exact routing.

Within the broader context of this study, this finding provides a crucial
structural counterpart to the binary addition model. While the adder
demonstrated the network's capacity to construct and topologically shield a
zero-error, infinite-horizon memory state (the continuous carry bit), the
multiplication controller illustrates the exact mathematical opposite. It
proves that the recurrent neuro-symbolic architecture can seamlessly
collapse into a pure, memoryless instruction decoder---analogous to a
deterministic hardware control unit---when temporal hysteresis is
algorithmically superfluous. Together, these dual results formally
demonstrate that minimal recurrent neural networks dynamically adapt their
internal phase space to the intrinsic theoretical complexity of the
underlying problem, either forging robust topological memory or reducing to
pristine Boolean routing as dictated by algorithmic necessity.

\subsubsection*{The Simplest Exact Solution}

A trivial analytical solution is 
\begin{equation}
h_{1,t}=\tanh (2x_{t}-1),\qquad h_{2,t}=0,\qquad y_{\text{gate}%
,t}=h_{1,t},\qquad y_{\text{shift},t}=1,  \label{simp}
\end{equation}%
where the recurrent memory matrix $W_{hh}$ is effectively nullified. The
existence of such a configuration carries profound theoretical implications
for neural algorithmic synthesis and exposes an important divergence between
human cognitive paradigms and machine optimization.

From a human perspective, the most obvious and elegant exact solution is (%
\ref{simp}). Humans inherently rely on symmetry, zeroed matrices, and
algebraic simplicity as a necessary scaffolding to construct analytic proofs
and comprehend logic. However, during training, the RNN consistently
bypasses this human-centric simplicity, discovering exact solutions like Eq.
(\ref{solmult}) that appear algebraically messy and asymmetric to us. This
discrepancy is a significant strength of the neural architecture.

The continuous RNN is unbound by the human need for aesthetic mathematical
simplicity. Instead of seeking isolated, highly symmetric points, it
discovers an entirely different concept of \textquotedblleft
simplicity\textquotedblright\ driven by the thermodynamics of gradient
descent: it locates broad, topologically open subsets in the parameter space
that inherently absorb floating-point truncation noise. By leveraging the
topological properties of the invariant basins, logical regions and
latent-space dynamics, the network reaches infinite stability without
requiring the rigid algebraic simplifications that typically bottleneck
human mathematical construction.

Liberated from these anthropocentric constraints, the network's freedom
within the parameter space is vastly expanded. This topological flexibility
considerably augments the set of reachable exact solutions, transforming
what would otherwise be a mathematically improbable search into a reliable
descent toward a vast open domain of perfect algorithmic configurations.

\section{The Universal Topological Quantization Conjecture}

\label{conj}

The empirical results and analytical proofs of zero-error structural
stability for continuous binary addition and for the multiplication
controller extend far beyond the specific tasks of sequential arithmetic.
They reveal a profound capability of recurrent continuous dynamical systems:
when the optimization landscape is properly sculpted, neural networks do not
merely approximate target discrete functions, but spontaneously quantize
into exact, structurally stable logical topologies.

Building upon these results, we formulate the following \textit{Topological
Quantization Conjecture}:

\begin{enumerate}
\item \textbf{Universal Structural Embodiment}. Any exact discrete
algorithmic task can be structurally embodied by a continuous Recurrent
Neural Network without relying on infinite numerical precision,
discontinuous step activations, or irrational weights. Exact algebraic
configurations are naturally accessible to finite-precision continuous
dynamical systems.

\item \textbf{Topological Openness in Parameter Space}. For any such logical
algorithm, zero-error structurally stable configurations are not restricted
to isolated anomalies, but include topologically open, macroscopic subsets
within the continuous parameter space of the network.

\item \textbf{Phase Space Quantization and Autonomous Learnability}. These
robust parameter sets sculpt the continuous latent phase space, generating
disjoint, error-absorbing invariant basins of attraction that permanently
trap the dynamic trajectories within compact logical zones separated by
finite gaps and mapped into their interiors by the dynamics. Because
nontrivial open subsets in parameter space encompass a dense set of rational
configurations, continuous networks can perfectly lock into discrete logic
using standard finite-precision arithmetic. Consequently, optimized gradient
descent can spontaneously bypass continuous stagnation to autonomously
discover and populate these exact finite-state machines.
\end{enumerate}

If this conjecture holds true, the core challenge is to locate these
structurally stable invariant basins and logical zones. While saturated
penalties like the repulsive dead-zone loss provide the essential
thermodynamic pressure to force quantization, identifying these attractors
in highly sparse constraint spaces (such as the 16-addition dataset used for
Solution 7 of Appendix \ref{A} and the Distilled Model C of Appendix \ref{B}%
---see Subsection \ref{thermo} for details) requires rigorous algorithmic
intervention, bypassing gradient descent inertia through hybrid multi-cycle
explorations.

Our findings also suggest that to isolate pure algorithmic logic, both
excessively large latent spaces and massive training datasets are
unnecessary. In highly dimensional settings, the continuous dynamics possess
a high number of degrees of freedom to \textquotedblleft
cheat\textquotedblright\ via statistical approximations, drifting away from
exact quantization. By operating in minimal dimensional spaces, we eliminate
these continuous escape routes, forcing the geometry to structurally
fracture into the precise algebraic skeleton required to solve the task.
Similarly, vast amounts of training data are not needed to induce this phase
transition, as evidenced by our ability to reach exact zero-error
quantization for the binary adder with a dataset of just 16 examples.

\section{The Physics of Amplification and the Verticalization of Neural
Architectures}

\label{superAI}

The exact analytical solutions and phase transitions detailed in this work
reveal a profound, fundamental capability of Recurrent Neural Networks: the
potential to achieve exact, autonomous learning. When the optimization
landscape is properly sculpted---such as through the repulsive dead-zone
penalty---neural networks do not merely approximate target functions; they
spontaneously leap from continuous, probabilistic mapping into exact,
quantized logical structures. This demonstrates that perfect algorithmic
reasoning is mathematically accessible within the parameter space and can be
autonomously reached by optimized gradient descent without requiring
infinite precision.

This insight prompts a critical reevaluation of the current trajectory
toward Artificial General Intelligence (AGI). The prevailing AGI paradigm
relies heavily on scaling monolithic, black-box architectures, operating
under the assumption that a sufficiently large number of parameters will
eventually approximate any conceivable human behavior. However, this
brute-force approach is statistically inefficient and highly vulnerable to
hallucination, catastrophic forgetting and continuous noise accumulation.
Building a super-intelligent system capable of zero-error infinite-horizon
reasoning is a statistically insurmountable task if approached as a single,
monolithic optimization problem.

To overcome this, we propose a paradigmatic shift from horizontal scaling to
the verticalization of AI, governed by the principles of what we term the 
\textit{physics of amplification}. Drawing directly from the modular
composability of our hybrid multiplier---which combines a memory-preserving
topological adder with a memoryless logical controller---titanic
computational goals (i.e., tasks so statistically disfavored as to be
practically unattainable) must be decomposed into a hierarchically nested
architecture of micro-problems, such that, for each of them:

\begin{enumerate}
\item convergence to an exact solution is so statistically favored as to be
practically certain;

\item the solution is either infinitely stable or sufficiently robust to act
as an unshakeable foundation for all higher-level computational layers.
\end{enumerate}

Ultimately, the verticalization of AI into a theoretically unbounded
\textquotedblleft Super AI\textquotedblright\ does not lie in building a
single macroscopic black box encompassing larger approximate models, but in
architecting phase-transitioned, structurally stable neural modules. The
\textquotedblleft artificial brain\textquotedblright\ can discover optimal
decompositions and correct vertical arrangements autonomously, replicating
the same neural mechanism at higher meta-levels, thereby building a
multi-dimensional network of nested networks (a fractal architecture where
meta-controllers dynamically route execution across topologically protected,
low-level modules).

Operating through optimized gradient descent, this architecture\ solves each
isolated micro-task almost immediately, reliably securing a stable parameter
configuration via topological phase transitions. Once acquired, these
protected micro-modules are frozen and composed hierarchically into an
interconnected structure. The resulting supernetwork amplifies its
capabilities vertically, achieving infinite-horizon reasoning without
succumbing to high-dimensional optimization paralysis or numerical drift.

The empirical and theoretical results of this paper prove that such
autonomous verticalization is not merely a theoretical construct, but a
concrete computational reality. As demonstrated by our dual arithmetic
tasks, the optimization landscape acts as a geometric and algorithmic
Occam's razor, inherently learning with zero error what to preserve and what
to discard. When persistent memory is mandatory---as in the binary
adder---the continuous dynamics construct and topologically shield a
zero-error infinite-horizon carry state. Conversely, when temporal
hysteresis is algorithmically superfluous---as in the multiplication
controller---gradient descent autonomously prunes excess degrees of freedom
and suppresses unnecessary state accumulation, seamlessly collapsing into a
memoryless instruction decoder. Ultimately, through this dual capacity to
either forge robust memory or reduce to pristine routing, the system proves
its ability to dynamically reshape its internal phase space, perfectly
orchestrating continuous neural dynamics to achieve exact, complex reasoning.

\subsection{Scale-Invariant Topology of Nested Networks}

We summarize here the fractal structure of the nested networks mentioned
above. It describes a self-similar, multi-level hierarchy designed to
replace monolithic \textquotedblleft black-box\textquotedblright\ models
with a multi-dimensional network of phase-transitioned, structurally stable
neural modules.

\subsubsection*{1. The Layered Fractal Hierarchy}

Rather than forcing a single macroscopic network to approximate an entire
algorithmic mapping end-to-end, the system replicates the same neural
mechanism across higher meta-levels:

\begin{itemize}
\item Base Layer (Topologically Protected Micro-Modules / Neural Logic
Units):

At the foundational level reside minimal neural architectures (such as the $%
d=2,3$, and $4$ recurrent arithmetic modules of this paper). When trained
under the repulsive dead-zone loss, the underlying networks undergo an
abrupt topological phase transition that quantizes the continuous latent
space into structurally stable basins of attraction, strictly confining the
dynamic states within logical regions separated by a finite gap, mapped into
their interiors by the dynamics. Because of this structural isolation, the
corresponding space of weight configurations contains open subsets within
the parameter space. Once learning is complete---as evidenced by the abrupt
collapse of the loss function---these neural modules are frozen into
immutable logic units.

\item Middle Layer (Neural Controllers \& Execution Routing):\newline
Lightweight neural controllers operate above the frozen base layer. Instead
of relearning low-level micro-tasks, they act as dynamic routers that manage
temporal control flow and conditional gating across the frozen modules (as
demonstrated by the neuro-symbolic multiplier orchestrating a frozen adder).
Governed by the geometric pressure of the loss function, these controllers
dynamically adapt to the complexity of the routing problem: when temporal
hysteresis is not required, the optimization landscape acts as an
algorithmic Occam's razor, pruning excess degrees of freedom and collapsing
the network into a pure, memoryless instruction decoder.

\item Higher Meta-Layers (Multi-Dimensional Nested Lattices):\newline
This \textquotedblleft module + controller\textquotedblright\ pairing serves
as a fractal unit: once its internal routing logic is perfected, the entire
hybrid system is frozen to become a new foundational module.
Meta-controllers at higher abstraction levels treat these composite units as
operationally immutable building blocks, dynamically routing signals across
a three- or multi-dimensional network of nested networks.
\end{itemize}

Importantly, while these underlying composite modules remain strictly frozen
during the amplification process to guarantee zero-error execution, they are
not locked for eternity. Over extended operational timescales, they could be
selectively unfrozen\ to revisit, relearn, or overwrite obsolete logic,
perfectly mirroring long-term neuroplasticity.

\subsubsection*{2. Foundational Principles Governing the Architecture}

The nested fractal structure relies on three core principles:

\begin{enumerate}
\item Statistical Certainty of Micro-Tasks: Titanically complex or
statistically insurmountable tasks are fractured into elementary
sub-problems simple enough for optimized gradient descent to solve almost
immediately with near certainty.

\item Topological Protection and Structural Stability: To act as an
unshakeable foundation for all higher-level computational layers, the
solution must be either infinitely stable or sufficiently robust. Strict
infinite stability is achieved when each micro-task is quantized into
structurally stable basins of attraction that rigorously confine the dynamic
states within compact logical zones separated by a finite gap and mapped by
the dynamics into their interiors. This open-set property ensures that
internal states absorb single-precision (float32) truncation noise without
phase drift over arbitrary sequence lengths.

\item Hierarchical Composability: Frozen, noise-immune modules are composed
hierarchically without catastrophic forgetting. Higher-level networks
dynamically adapt their internal phase space---either forging topological
memory or collapsing into pure Boolean routing as dictated by algorithmic
necessity.
\end{enumerate}

\section{Conclusions}

\label{sec:conclusions}

We addressed the fundamental challenge of mechanistic interpretability in
Recurrent Neural Networks by proving that continuous dynamical systems do
not merely approximate target logic, but can spontaneously quantize into
exact, error-tolerant Boolean machines.

Understanding how a continuous, differentiable dynamical system can execute
exact symbolic algorithms over unbounded time horizons requires looking
beyond standard statistical learning paradigms. As demonstrated throughout
this work, the inability of conventional RNNs to crystallize into exact
finite-state machines is not an architectural limitation of continuous
activation functions such as the hyperbolic tangent, but a pathology of
standard optimization objectives. Under conventional Cross-Entropy loss,
non-vanishing gradients relentlessly push well-classified states toward
asymptotic saturation, inflating weight magnitudes and producing brittle,
chaotic decision boundaries that inevitably fail under floating-point
truncation noise. By introducing the repulsive dead-zone loss $\mathcal{L}_{%
\text{RDZ}}$, we fundamentally alter these optimization thermodynamics.
Combining a zero-gradient neutral zone at high confidence with an actively
repulsive logarithmic barrier against extreme saturation, $\mathcal{L}_{%
\text{RDZ}}$ prevents geometric degeneration and escapes stagnation, forcing
the continuous phase space to undergo an abrupt topological phase transition
to Boolean logic, which locks the system into globally stable geometric
attractors.

The mathematical core of this transition lies in the global bifurcation of
the latent manifold. Rather than relying on the unphysical assumptions of
classical automata theory---such as discontinuous step activations, infinite
weight limits, or infinite-precision irrational parameters---the networks
analyzed here fracture their smooth state space into disjoint, invariant
memory basins $I_{0}$ and $I_{1}$ that enclose compact logical regions $%
A_{ij}$ which are separated by finite gaps and mapped into their interiors
by the dynamics in latent space.

In the minimal latent dimension $d=2$, this quantization emerges
spontaneously during training even on highly sparse datasets of just 16
binary additions, yielding empirical configurations that admit an exact
rank-1 dimensional reduction onto a one-dimensional scalar subspace.

Extending the analysis to $d=3$ and $d=4$ further clarifies the relationship
between algebraic symmetry and dynamical robustness: while human-crafted
closed-form solutions naturally favor symmetric weight configurations, such
arrangements typically sit on fragile, zero-gap boundaries that leave the
system marginally stable and vulnerable to asymptotic decay. Breaking these
symmetries by algebraically twisting specific parameters, we eliminated
configurations lacking strict geometric separation, structurally isolating
the logical regions by finite margins.

In all cases, crucial properties---such as the finite separation between the
compact logical regions $A_{ij}$ and the fact that the dynamics maps them
into their interiors---prove that the exact, infinitely stable solutions are
not isolated mathematical points, but include nonempty open subsets in
parameter space encompassing a dense set of dyadic rationals. This
topological openness reconciles continuous dynamics with finite-precision
hardware and guarantees that an appropriately optimized RNN inevitably
converges into logical regions formed by error-absorbing memory and readout
basins, without requiring infinite precision, provided it operates on a
suitable discrete arithmetic.

The required numerical tolerance is dictated by the minimum geometric safety
margin of the architecture---encompassing both the gap between the logical
regions $A_{ij}$ and the internal distances separating the mapped images $%
H(S,A_{ij})$ from the $A_{ij}$ boundaries. In the architectures considered
in this paper, this safety margin is sufficiently wide that standard float32
representation is amply sufficient to prevent any numerical drift.

We extended the framework to multi-bit binary multiplication by freezing the
exact neural adder as an immutable ALU directed by a trainable recurrent
controller. A zero-error hybrid multiplier orchestrating sequential
shift-and-add logic was obtained, revealing that under the geometric
pressure of the repulsive dead-zone loss, optimization acts as an
algorithmic Occam's razor: whereas the adder forges a persistent hysteresis
loop to propagate the carry bit, the controller autonomously prunes
redundant degrees of freedom to operate as a memoryless combinatorial
router. The solution highlights that the optimization process is unbound by
human aesthetic biases. Unconstrained by the anthropocentric search for
symmetry, the network explores a vastly magnified parameter space.
Stochastic descent naturally converges into broad, asymmetric open basins
that physiologically absorb truncation noise, guaranteeing absolute zero
error over arbitrary sequence lengths without accumulating floating-point
drift.

Together with the observation that minimal latent dimensions and small
training sets eliminate continuous escape routes---preventing the network
from cheating via statistical interpolation---this structural duality
between the synthesis of persistent topological memory and the collapse into
memoryless Boolean routing underpins the Universal Topological Quantization
Conjecture: any discrete algorithmic task is naturally embodied by open,
rational-dense regions within the parameter space of a minimal continuous
network.

In a broader perspective, these results transcend the analysis of individual
arithmetic tasks to challenge the prevailing AI paradigm of flat, monolithic
horizontal scaling. Expanding black-box architectures end-to-end remains
statistically inefficient and structurally vulnerable to hallucination,
phase drift, and catastrophic forgetting.

By contrast, the physics of amplification outlined in this paper
demonstrates that seemingly unattainable computational goals can be
fractured into a vertical, multi-dimensional fractal hierarchy of minimal,
statistically certain, and infinitely stable micro-modules. Because each of
these units can be driven with near certainty into an open, error-absorbing
geometric attractor and subsequently frozen, higher-level meta-controllers
can orchestrate complex execution flows over a solid logical scaffolding,
establishing a mathematically rigorous, noise-immune foundation toward a
theoretically unbounded \textquotedblleft Super AI\textquotedblright\
capable of verifiable, infinite-horizon reasoning.

\vskip1truecm

{\textbf{\huge Appendices}} \renewcommand{\thesection}{\Alph{section}} %
\renewcommand{\theequation}{\thesection.\arabic{equation}} %
\setcounter{section}{0}

\section{Empirically Determined Weight Tensors}

\label{A}\setcounter{equation}{0}

This Appendix compiles several network parameter sets obtained via
stochastic gradient descent under repulsive dead-zone loss regularization in 
$d=2$. The size of the dataset used for training is specified in parentheses
for each model.

The majority of the discovered systems constitute isomorphic configurations
of the same underlying phase-space attractors (up to trivial symmetry
operations, such as permutations of the latent coordinates $h_{1}$ and $%
h_{2} $, as well as spatial reflections). Solutions 1-6, trained on the 256-
and 64-addition datasets, appear to belong to the same class, while Solution
7, trained on the 16-addition dataset, appears to belong to a different
class.

In all cases, the network autonomously recovered the commutative property of
binary addition with respect to $x_{1}$ and $x_{2}$, as evidenced by all
input matrices $W_{ih}$ consistently exhibiting two practically identical
columns.

We evaluated the RNN predictions of each solution against exact Boolean
logic across comprehensive stress tests. These included: all possible sums
of binary numbers up to and including 14 bits; $10^{5}$ random sequences of
250 bits; 10$^{4}$ random sequences of 10$^{3}$ bits; and adversarial edge
cases (pure cascade, double cascade, delayed carry, unbalanced, and
high-density numbers) extending to lengths of 100, 500, 10$^{3}$, 5$\cdot $10%
$^{3}$, 10$^{4}$, 10$^{5}$, and 10$^{6}$ bits. No computational failures
were ever detected, providing strong empirical evidence of zero-error
infinite stability.

Solution 1 (dataset 256)%
\begin{eqnarray*}
W_{ih} &=&%
\begin{pmatrix}
2.2780404 & \text{ }2.2746418 \\ 
0.2732029 & \text{ }0.27182734%
\end{pmatrix}%
,\qquad W_{hh}=%
\begin{pmatrix}
2.163378 & \text{ }-3.46534 \\ 
0.25362742 & \text{ }-0.40067995%
\end{pmatrix}%
, \\
W_{fc} &=&%
\begin{pmatrix}
4.252388 & \text{ }-16.640718 \\ 
-4.5407777 & 16.66741%
\end{pmatrix}%
,\qquad 
\begin{tabular}{l}
$b_{h}=%
\begin{pmatrix}
-3.4846325 & \text{ }-0.6773404%
\end{pmatrix}%
^{T},$ \\ 
$b_{fc}=%
\begin{pmatrix}
-4.3212404 & \text{ }4.321245%
\end{pmatrix}%
^{T}.$%
\end{tabular}%
\end{eqnarray*}

Solution 2 (dataset 256)%
\begin{eqnarray*}
W_{ih} &=&%
\begin{pmatrix}
-2.2462764 & \text{ }-2.246856 \\ 
-0.26404023 & \text{ }-0.26408726%
\end{pmatrix}%
,\qquad W_{hh}=%
\begin{pmatrix}
2.122489 & \text{ }-3.474678 \\ 
0.2493516 & \text{ }-0.40081894%
\end{pmatrix}%
, \\
W_{fc} &=&%
\begin{pmatrix}
-4.393065 & \text{ }16.567516 \\ 
4.136624 & \text{ }-16.502674%
\end{pmatrix}%
,\qquad 
\begin{tabular}{l}
$b_{h}=%
\begin{pmatrix}
3.4273505 & \text{ }0.66880965%
\end{pmatrix}%
^{T},$ \\ 
$b_{fc}=%
\begin{pmatrix}
-4.303299 & \text{ }4.303301%
\end{pmatrix}%
^{T}.$%
\end{tabular}%
\end{eqnarray*}

Solution 3 (dataset 256)%
\begin{eqnarray*}
W_{ih} &=&%
\begin{pmatrix}
2.2903519 & \text{ }2.2888908 \\ 
0.27264705 & \text{ }0.27208534%
\end{pmatrix}%
,\qquad W_{hh}=%
\begin{pmatrix}
2.1687353 & \text{ }-3.458624 \\ 
0.2535042 & \text{ }-0.3980963%
\end{pmatrix}%
, \\
W_{fc} &=&%
\begin{pmatrix}
4.4713764 & \text{ }-16.702353 \\ 
-4.3072896 & \text{ }16.485281%
\end{pmatrix}%
,\qquad 
\begin{tabular}{l}
$b_{h}=%
\begin{pmatrix}
-3.4887831 & \text{ }-0.67768496%
\end{pmatrix}%
^{T},$ \\ 
$b_{fc}=%
\begin{pmatrix}
-4.289307 & \text{ }4.2893047%
\end{pmatrix}%
^{T}.$%
\end{tabular}%
\end{eqnarray*}

Solution 4 (dataset 256)%
\begin{eqnarray*}
W_{ih} &=&%
\begin{pmatrix}
2.2989724 & \text{ }2.3004158 \\ 
-0.27625507 & \text{ }-0.2757567%
\end{pmatrix}%
,\qquad W_{hh}=%
\begin{pmatrix}
2.1935937 & \text{ }3.4443886 \\ 
-0.25828683 & \text{ }-0.39784217%
\end{pmatrix}%
, \\
W_{fc} &=&%
\begin{pmatrix}
4.406853 & \text{ }16.504414 \\ 
-4.4391336 & \text{ }-16.41644%
\end{pmatrix}%
,\qquad 
\begin{tabular}{l}
$b_{h}=%
\begin{pmatrix}
-3.496089 & \text{ }0.68003047%
\end{pmatrix}%
^{T},$ \\ 
$b_{fc}=%
\begin{pmatrix}
-4.21858 & \text{ }4.2185783%
\end{pmatrix}%
^{T}.$%
\end{tabular}%
\end{eqnarray*}

Solution 5 (dataset 256)%
\begin{eqnarray*}
W_{ih} &=&%
\begin{pmatrix}
0.2741506 & \text{ }0.2749987 \\ 
-2.275918 & \text{ }-2.278653%
\end{pmatrix}%
,\qquad W_{hh}=%
\begin{pmatrix}
-0.40181336 & \text{ }-0.25755996 \\ 
3.46777 & \text{ }2.1874232%
\end{pmatrix}%
, \\
W_{fc} &=&%
\begin{pmatrix}
-16.387323 & \text{ }-4.4172416 \\ 
16.653149 & \text{ }4.4256606%
\end{pmatrix}%
,\qquad 
\begin{tabular}{l}
$b_{h}=%
\begin{pmatrix}
-0.6787486 & \text{ }3.4728334%
\end{pmatrix}%
^{T},$ \\ 
$b_{fc}=%
\begin{pmatrix}
-4.238094 & \text{ }4.238087%
\end{pmatrix}%
^{T}.$%
\end{tabular}%
\end{eqnarray*}

Solution 6 (dataset 64) 
\begin{eqnarray*}
W_{ih} &=&%
\begin{pmatrix}
-0.30071792 & \text{ }-0.30039257 \\ 
-2.1383412 & \text{ }-2.1402447%
\end{pmatrix}%
,\qquad W_{hh}=%
\begin{pmatrix}
-0.45960948 & \text{ }0.30744687 \\ 
-3.4631953 & \text{ }2.2295809%
\end{pmatrix}%
, \\
W_{fc} &=&%
\begin{pmatrix}
15.993112 & \text{ }-4.6905174 \\ 
-16.018242 & \text{ }4.8296156%
\end{pmatrix}%
,\qquad 
\begin{tabular}{l}
$b_{h}=%
\begin{pmatrix}
0.72108346 & \text{ }3.3032732%
\end{pmatrix}%
^{T},$ \\ 
$b_{fc}=%
\begin{pmatrix}
-4.033971 & \text{ }4.0339713%
\end{pmatrix}%
^{T}.$%
\end{tabular}%
\end{eqnarray*}

Solution 7 (dataset 16) 
\begin{eqnarray*}
W_{ih} &=&%
\begin{pmatrix}
0.14678793 & \text{ }0.14679323 \\ 
6.3393297 & \text{ }6.251233%
\end{pmatrix}%
,\qquad W_{hh}=%
\begin{pmatrix}
0.26918396 & \text{ }0.04202239 \\ 
10.597283 & \text{ }2.130171%
\end{pmatrix}%
, \\
W_{fc} &=&%
\begin{pmatrix}
-20.42872 & \text{ }2.655774 \\ 
21.148771 & \text{ }-3.2674134%
\end{pmatrix}%
,\qquad 
\begin{tabular}{l}
$b_{h}=%
\begin{pmatrix}
0.0065637077 & \text{ }-8.596262%
\end{pmatrix}%
^{T},$ \\ 
$b_{fc}=%
\begin{pmatrix}
4.6880956 & \text{ }-3.7119586%
\end{pmatrix}%
^{T}.$%
\end{tabular}%
\end{eqnarray*}

Analogous empirical results (not detailed here) were obtained by training
RNNs in latent-space dimensions $d=3$ and $d=4$.

\section{Distilled Exact Configurations}

\label{B}\setcounter{equation}{0}

In this Appendix we present polished models inspired by the RNN\ solutions
listed above. All passed the stress tests mentioned in the previous Appendix.

\vskip .3truecm

\textbf{Distilled Model A}

\vskip .1truecm

This configuration was derived from the empirical RNN Solution 3, with the
primary objective of simplifying the numerical parameters: 
\begin{eqnarray*}
W_{ih} &=&%
\begin{pmatrix}
2.25 & \text{ }2.25 \\ 
0.25 & \text{ }0.25%
\end{pmatrix}%
,\qquad W_{hh}=%
\begin{pmatrix}
2.25 & \text{ }-3.5 \\ 
0.25 & \text{ }-0.4%
\end{pmatrix}%
, \\
W_{fc} &=&%
\begin{pmatrix}
0.0 & \text{ }0.0 \\ 
-1.0 & \text{ }4.0%
\end{pmatrix}%
,\qquad 
\begin{tabular}{l}
$b_{h}=%
\begin{pmatrix}
-3.5 & \text{ }-0.7%
\end{pmatrix}%
^{T},$ \\ 
$b_{fc}=%
\begin{pmatrix}
0.0 & \text{ }1.0%
\end{pmatrix}%
^{T}.$%
\end{tabular}%
\end{eqnarray*}

Arbitrary manual perturbations generally compromise the model's integrity,
causing it to fail the stress tests. The successful extraction of distilled
models demonstrates that the subspace of infinitely stable solutions
includes an open set within the parameter space, corroborating the graphical
and analytical proofs presented in Sections \ref{sec:empirical_verification}
and \ref{sec:invariance}.

\vskip .3truecm

\textbf{Distilled Model B}

\vskip .1truecm

This is the arrangement used for the proofs of Sections \ref%
{sec:empirical_verification} and \ref{sec:invariance}, suggested by the RNN\
Solution 5:%
\begin{eqnarray*}
W_{ih} &=&%
\begin{pmatrix}
0.25 & \text{ }0.25 \\ 
-2.25 & \text{ }-2.25%
\end{pmatrix}%
,\qquad W_{hh}=%
\begin{pmatrix}
-0.4 & \text{ }-0.25 \\ 
3.5 & \text{ }2.1875%
\end{pmatrix}%
, \\
W_{fc} &=&%
\begin{pmatrix}
0.0 & \text{ }0.0 \\ 
4.0 & \text{ }1.0%
\end{pmatrix}%
,\qquad 
\begin{tabular}{l}
$b_{h}=%
\begin{pmatrix}
-0.7 & \text{ }3.5%
\end{pmatrix}%
^{T},$ \\ 
$b_{fc}=%
\begin{pmatrix}
0.0 & \text{ }1.0%
\end{pmatrix}%
^{T}.$%
\end{tabular}%
\end{eqnarray*}%
With respect to the Distilled Model A, it entails trivial operations (a
permutation of $h_{1}$ and $h_{2}$, and a reflection along the new $h_{1}$),
but also, and most importantly, a change of the $W_{hh}$ entry from $2.25$
into $2.1875$. This additional modification was deliberately sought to
render the matrix $W_{hh}$ singular, reducing its rank to one and thereby
lowering the dimensionality of the problem: the evolution $%
h_{t-1}\rightarrow h_{t}$ maps the square $[-1,1]^{2}$ into a curve for
every $S_{t}$, making the graphical proof of Section \ref%
{sec:empirical_verification} and the analytic proof of Section \ref%
{sec:invariance} relatively straightforward.

It took a few trials to identify an empirically stable arrangement with this
key property. Attempts to achieve the same reduction by modifying the first
row of $W_{hh}$ instead of the second one failed. 
\begin{figure}[tb]
\centering\includegraphics[width=16truecm]{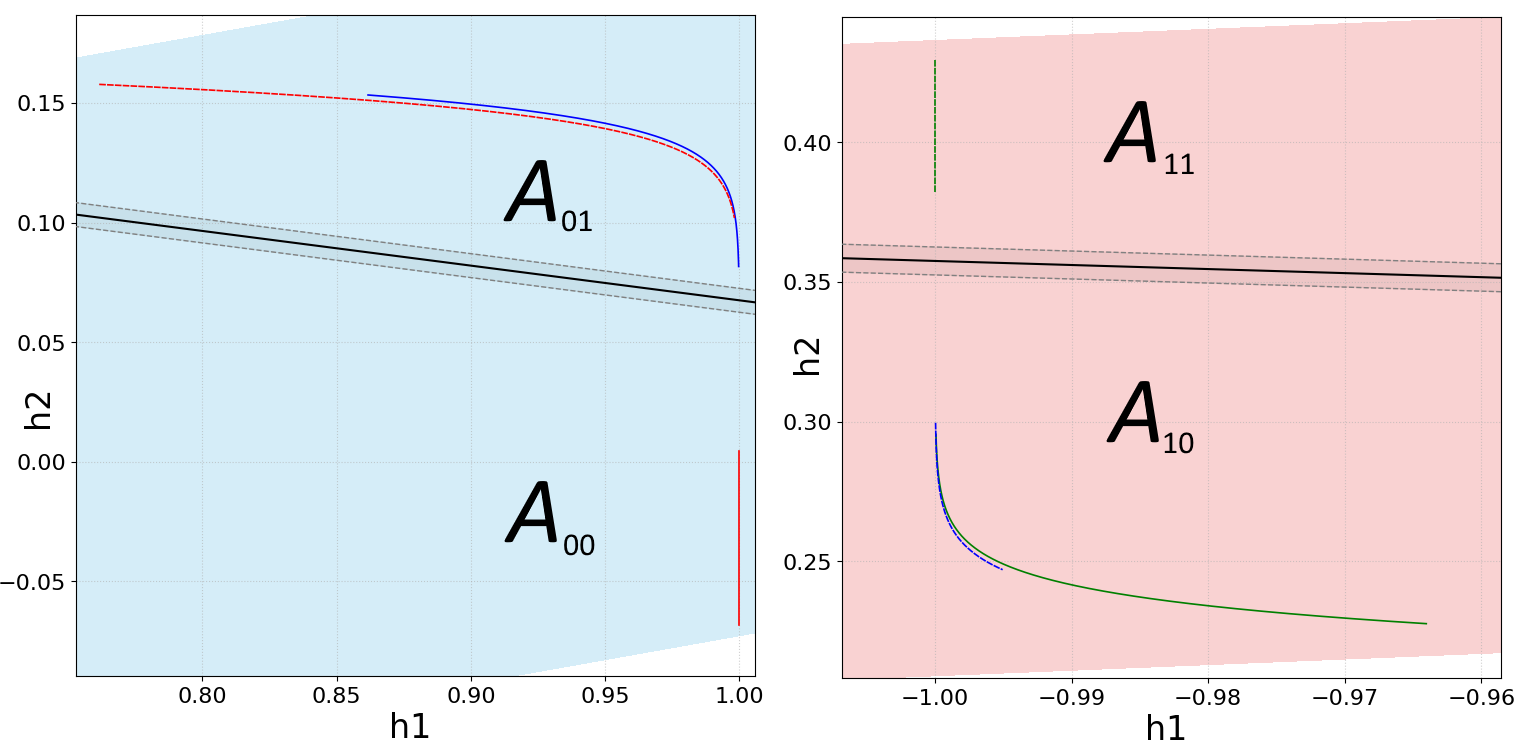}
\caption{Proof of zero-error infinite stability for Distilled Model C}
\label{graph_b_16}
\end{figure}

\vskip .3truecm

\textbf{Distilled Model C}

\vskip .1truecm

Inspired by a variant of Solution 7, also trained on the 16-addition
dataset, we derived a second instance of dimensional reduction with a rank-1
recurrent matrix $W_{hh}$: 
\begin{eqnarray*}
W_{ih} &=&%
\begin{pmatrix}
-6.5 & \text{ }-6.5 \\ 
0.15 & \text{ }0.15%
\end{pmatrix}%
,\qquad W_{hh}=%
\begin{pmatrix}
2.2 & \text{ }-11.0 \\ 
-0.05 & \text{ }0.25%
\end{pmatrix}%
, \\
W_{fc} &=&%
\begin{pmatrix}
-2.8 & \text{ }-20.0 \\ 
3.0 & \text{ }20.0%
\end{pmatrix}%
,\qquad 
\begin{tabular}{l}
$b_{h}=%
\begin{pmatrix}
8.0 & \text{ }0.0%
\end{pmatrix}%
^{T},$ \\ 
$b_{fc}=%
\begin{pmatrix}
4.5 & \text{ }-4.0%
\end{pmatrix}%
^{T}.$%
\end{tabular}%
\end{eqnarray*}%
While this configuration still confines the memory evolution to a
one-dimensional scalar bottleneck, it operates through an asymmetric
parameter scaling rather than a balanced distribution. Although the latent
variables remain functionally coupled to correctly route the carry signal,
this scale asymmetry forces the two coordinates to operate in fundamentally
different non-linear regimes of the hyperbolic tangent.

The graphical proof of zero-error infinite stability for this model is
presented in Fig. \ref{graph_b_16}. The corresponding phase space mappings
can be interpreted following the same topological methodology detailed for
Distilled Model B in Section \ref{sec:empirical_verification}. The scalar
compression variable $Z$ and the output variable $Y$, derived from the logit
difference $y_{t}=y_{2,t}-y_{1,t}$, are defined as%
\begin{equation*}
Z=2.2h_{1}-11h_{2},\qquad Y=5.8h_{1}+40h_{2}-8.5,
\end{equation*}%
while the memory basins $I_{0}$ and $I_{1}$ along $Z$ and the output basins $%
J_{0}$ and $J_{1}$ for $Y$ are given by 
\begin{equation}
I_{0}=[-0.2,3.0],\qquad I_{1}=[-7.0,-4.5]\text{,\qquad }J_{0}=[-54.3,-0.2],%
\qquad J_{1}=[0.2,37.3].  \notag
\end{equation}

\end{document}